\documentclass[lettersize,journal]{IEEEtran}
\usepackage{amsmath,amsfonts}
\usepackage{algorithmic}
\usepackage{algorithm}
\usepackage{array}
\usepackage[caption=false,font=normalsize,labelfont=sf,textfont=sf]{subfig}
\usepackage{textcomp}
\usepackage{stfloats}
\usepackage{url}
\usepackage{verbatim}
\usepackage{graphicx}
\usepackage{cite}
\usepackage[normalem]{ulem}
\usepackage[colorlinks=true,linkcolor=black,urlcolor=black,citecolor=black]{hyperref}
\usepackage{booktabs}
\usepackage{colortbl}
\usepackage{multirow}
\usepackage{multicol}
\usepackage{float}
\usepackage{soul}
\usepackage{bm}

\newtheorem{hypo}{Hypothesis}

\newcommand{\mc}[1]{\mathcal{#1}}

\def\eg{\emph{e.g.}}

\def\ie{\emph{i.e.}}

\begin{document}

\title{Hiding Faces in Plain Sight: Defending DeepFakes by Disrupting Face Detection}


\author{Delong Zhu, Yuezun Li, Baoyuan Wu, Jiaran Zhou, Zhibo Wang, Siwei Lyu, \IEEEmembership{Fellow,~IEEE}
\thanks{Yuezun Li is the {\em corresponding author}.}
\thanks{Delong Zhu, Yuezun Li, and Jiaran Zhou are with the School of Computer Science and Technology, Ocean University of China, Qingdao, China. Email: \{zhudelong\}@stu.ouc.edu.cn;\{liyuezun;zhoujiaran\}@ouc.edu.cn.}
\thanks{Baoyuan Wu is with the School of Data Science, The Chinese University of Hong Kong, Shenzhen, Guangdong, 518172, P.R. China. Email: wubaoyuan@cuhk.edu.cn}
\thanks{Zhibo Wang is with the School of Cyber Science and Technology, Zhejiang University, China, and ZJU-Hangzhou Global Scientific and Technological Innovation Center. Email: zhibowang@zju.edu.cn}
\thanks{Siwei Lyu is with the University at Buffalo, SUNY, USA. Email: siweilyu@buffalo.edu.}}

\markboth{Journal of \LaTeX\ Class Files,~Vol.~14, No.~8, August~2021}%
{Shell \MakeLowercase{\textit{et al.}}: A Sample Article Using IEEEtran.cls for IEEE Journals}


\maketitle

\begin{abstract}
Face-swapping DeepFakes have become an escalating societal concern, attracting increasing attention in recent years. To counter this, we investigate a new proactive defense framework to prevent individuals from being victimized in DeepFake videos. The core idea of this framework is to contaminate the inputs of DeepFake models by disrupting face detectors, based on the observation that face detectors are commonly used to automatically extract victim faces in most DeepFake techniques. Once the face detectors malfunction, the faces will not be correctly extracted, thereby impairing the training or synthesis stages of DeepFake models. To achieve this, we describe a strategy named {\em FacePoison}, which fools face detectors by adding dedicated adversarial perturbations to video frames. Building upon this, we introduce {\em VideoFacePoison}, an extended strategy that can efficiently propagate FacePoison across video frames instead of applying it individually to each frame, thus significantly reducing the computational overhead while retaining favorable attack performance. This framework is validated on five face detectors, and extensive experiments against eleven different DeepFake models demonstrate the effectiveness of disrupting face detectors to hinder DeepFake generation. The source code is publicly available at: \url{https://github.com/OUC-VAS/FacePoison}.
\end{abstract}


\section{Introduction}
\begin{figure}[t]
    \centering
    \includegraphics[width=\linewidth]{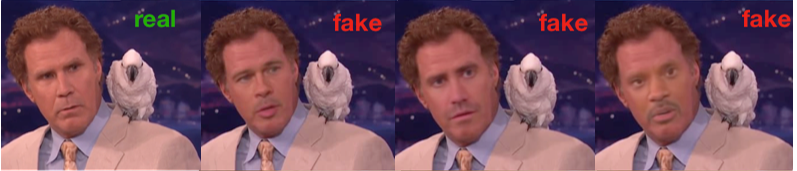}
    \vspace{-0.7cm}
    \caption{\small Examples of DeepFake, which involves replacing the original faces with synthesized faces while keeping the same facial expressions. These examples are from \cite{li2020celeb}.  }
    \label{fig:dfdemo}
\end{figure}

    \IEEEPARstart{R}{ecent} advances in deep learning and the availability of a vast volume of online personal images and videos have drastically improved the synthesis of highly realistic human faces ~\cite{karras2018progressive,karras2019style,karras2021alias}. DeepFake is one of the most prevalent face forgery techniques that has drawn increasing attention recently \cite{faceswap,liu2023deepfacelab,chen2020simswap,Gao_2021_CVPR,xu2022mobilefaceswap,shiohara2023blendface}\footnote{In this paper, the term DeepFake specifically refers to faces-swapping face forgery techniques, as originally defined in~\cite{spivak2018deepfakes}, where specific local facial content is replaced by newly synthesized content while retaining high visual quality. It is worth noting that the scope of DeepFake has recently expanded to encompass a wide range of AI-generated media, including full image synthesis and face editing techniques~\cite{luo2024diff,epstein2023diffusion,gong2023toontalker,yang2024learning,Huang_2023_CVPR,li2024strokefacenerf}. Considering that face-swapping DeepFakes remain the most prevalent form encountered in online content, our work focuses on this category.}. 
Primarily, DeepFake can swap the face with a synthesized target face while retaining the same facial attributes such as facial expression and orientation (Fig.~\ref{fig:dfdemo}). While there are interesting and creative applications of DeepFakes, they can also be weaponized to create illusions of a person's presence and activities that do not occur in reality, leading to serious political, social, financial, and legal consequences \cite{busacca2023deepfake,tahraoui2023defending}.

\begin{figure*}[t]
\centering
\includegraphics[width=\linewidth]{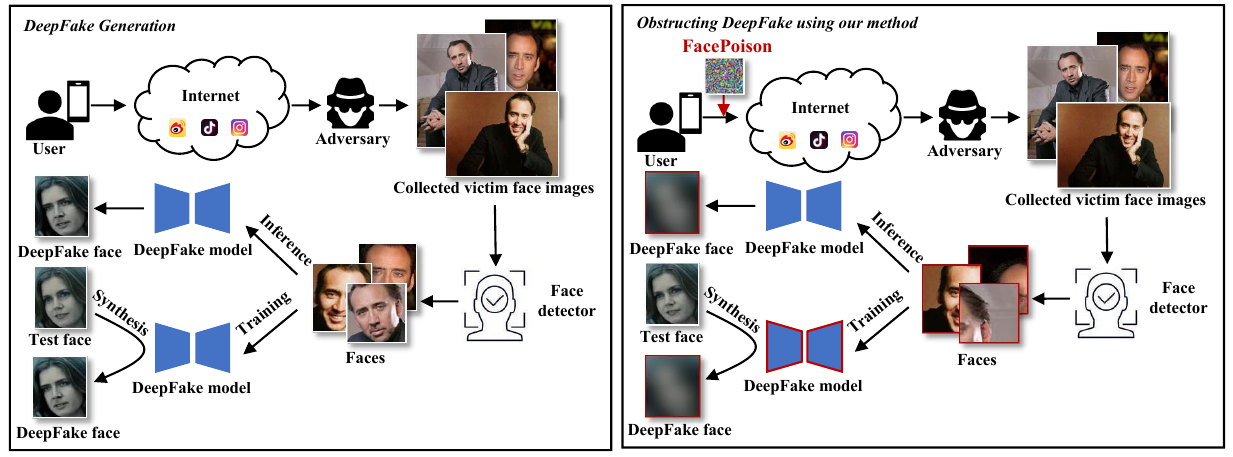}
\vspace{-0.7cm}
\caption{\small Overview of FacePoison. The left part shows the typical generation process of DeepFake conducted by adversaries, covering the inference and training phase of DeepFake models. The right part shows how our method obstructs the DeepFake generation. The rationale is that our method disrupts face detection, leading to incorrect face detection results. It can pollute the input faces during either inference or training, ultimately hindering the DeepFake generation process.}
\label{fig:overview}
\end{figure*}

Foreseeing this threat, many forensic methods aiming to detect DeepFake faces have been proposed recently \cite{lsda_2024_CVPR,li2018ictu,li2019exposing,diff-id,fei2022learning,chen2022selfsupervised,wang2023dynamic,ba2024exposing,cui2024forensics}. However, given the speed and reach of the propagation of online media, even the currently best forensic method will largely operate in a postmortem fashion, applicable only after the fake face images or videos emerge. In this work, we aim to develop {\em proactive} approaches to protect individuals from becoming the victims of such attacks. Our solution is to add specially designed patterns known as {adversarial perturbations} that are imperceptible to human eyes but can result in face detection failures. 
The key idea is that: \textit{DeepFake models, whether during training or inference, need many standard faces, \ie, face sets, collected using automatic face detection methods. In the training phase, a large number of standard faces are required as training sets. In the inference phase, standard faces also need to be extracted from each frame of the input videos. Our method aims to ``pollute'' this face set to disrupt the corresponding process} (see Fig. \ref{fig:overview}).

Our study, dubbed as {\em FacePoison}, focuses on adversarial attacks to the deep neural network (DNN)-based face detectors \eg, \cite{zhang2017s3fd,tang2018pyramidbox}, which have demonstrated superior performance over non-DNN approaches and exhibit greater robustness to variations in pose, expression, and occlusion. Specifically, we adapt mainstream adversarial attack methods for this task, with dedicated modifications on objectives, attack places, and attack processes. We comprehensively evaluate their effectiveness in disrupting face detection and the defense ability against DeepFake models. This study has practical implications and could be deployed as a privacy-preserving service on social media platforms, or as a standalone tool for users to preprocess their images before uploading them online.


Moreover, considering that users frequently upload videos to social platforms, we explore a strategy called {\em VideoFacePoison}, an extension of FacePoison designed for effective video protection. Since the adjacent frames in a video have strong temporal connections, we hypothesize that the adversarial perturbations across these frames also share temporal consistency. Based on this insight, we develop an optical flow-based strategy to propagate adversarial perturbations from one frame to adjacent frames, instead of calculating FacePoison on all frames.

Extensive experimental evaluations are performed on five mainstream face detectors on public face detection datasets, showing the effectiveness of our method in disturbing face detectors. We then apply our method to obstruct eleven well-known DeepFake models. The results demonstrate that our method can effectively degrade the visual quality of DeepFake faces. 

{\em It is important to highlight that the proposed framework is not a replacement but a complement to existing DeepFake detection methods.} The contribution of this paper is summarized as follows: 
\begin{enumerate}
    \item We describe a new proactive defense framework called {\em FacePoison} to obstruct DeepFake generation by disrupting DNN-based face detectors.

    \item We tailor and apply several mainstream adversarial attack techniques to this context, incorporating task-specific modifications, and thoroughly assess their effectiveness under various conditions.
    
    \item In light of the widespread sharing of personal videos online, we describe {\em VideoFacePoison}, which leverages temporal coherence within videos to propagate the FacePoison from a single frame to adjacent ones. Compared to applying FacePoison to all frames, VideoFacePoison offers an option to reduce computational overhead while retaining favorable attack performance.

    \item We conduct extensive experiments on five mainstream DNN-based face detection methods and eleven DeepFake models, demonstrating the effectiveness of our method in obstructing DeepFake generation in both training and testing. 
\end{enumerate}

This work extends our preliminary study presented in the {\tt ICME conference paper \cite{lifacepoison2023}} in the following aspects: 1) We adapt multiple mainstream adversarial attack methods to this task and comprehensively evaluate their feasibility and effectiveness; 2) We introduce VideoFacePoison, utilizing a novel optical flow-based strategy to propagate FacePoison across video frames; 3) We expand our experiments to include eleven well-known DeepFake models, covering defense scenarios in both training and testing phases.


\section{Background and Related Works}
\label{sec:background}

\subsection{DeepFake} 
DeepFake is a recent AI-based face forgery technique that can be used to create realistic impersonation videos. There are many variants of DeepFake versions to date, \eg, using different encoders, different decoders, or different loss functions \cite{fakeapp,DFaker,faceswap-gan,faceswap,liu2023deepfacelab}. These techniques share a common pipeline: Given an original video, the faces of source individuals are first cropped out using face detectors. Then these cropped faces are sent into a DeepFake model (\eg, auto-encoders~\cite{zhu2017toward}, GANs~\cite{goodfellow2014generative} or Diffusion models~\cite{ho2020denoising}) to synthesize new blended faces. These synthesized faces belong to the source individuals but have the same facial attributes, such as facial expressions and head orientations, as in the original video. 
Nowadays, with the thriving of generative models, DeepFake has not been limited to the technique of face-swap and has become a general term for all AI-based face forgery techniques, \eg, face editing \cite{li2024strokefacenerf}, full face generation \cite{karras2019style}, face inpainting \cite{sola2023unmasking}, etc. This paper still focuses on defending the face-swap DeepFakes. 

\subsection{Face Detection} 
Using DNNs in face detection has become mainstream with their high performance and robustness regarding variations in pose, expression, and occlusion. There has been a plethora of DNN-based face detectors \eg, \cite{zhang2017s3fd,deng2020retinaface, qi2022yolo5face,tang2018pyramidbox,li2019dsfd}. Regardless of the idiosyncrasies of different detectors, they all follow a similar workflow, which predicts the location and confidence score of the potential candidate regions corresponding to faces in an end-to-end fashion. 
The prohibitive cost of searching optimal network structures and architectures makes the choice of the backbone network limited to four well-tested DNN models, namely, the VGG network \cite{simonyan2014very}, ResNet \cite{he2016deep}, MobileNet \cite{howard2017mobilenets}, and ShuffleNet \cite{zhang2018shufflenet} as reported on the leader board of the WIDER challenge \cite{yang2016wider}. 

\subsection{Adversarial Perturbations} 
\label{sec:adv}
Adversarial Perturbations are intentionally designed noises that are imperceptible to human observers yet can seriously reduce the performance of deep neural networks when added to the input image. Many methods \cite{goodfellow2014explaining,kurakin2016adversarial,moosavi2016deepfool,dong2018boosting,9453106,9207872,8781934,10163476,xie2019improving,lin2019nesterov} have been proposed to impair image classifiers by adding adversarial perturbations on the entire image. The following briefly overviews several classic attack methods relevant to this work.

\smallskip
\noindent{\bf FGSM.} 
Fast Gradient Sign Method (FGSM) \cite{goodfellow2014explaining} is the method that can update the input image within a distortion bound in one step based on the sign of the gradient, calculated by a designed objective. Let $x,y$ be the input image and its ground-truth label, and $x^{adv}$ be the generated adversarial image. The update process to obtain $x^{adv}$ can be written as
\begin{equation}
	x^{adv} = x + \alpha \cdot sign(\nabla_{x}\mc{L}(x, y)), \; s.t. \; || x^{adv} - x ||_{\infty} \leq \epsilon,
\end{equation}
where $\mc{L}(\cdot, \cdot)$ is the designed objective function, $\nabla_{x}$ is the gradient with respect to image $x$, $\alpha$ is the step size and $sign$ is a function that retains the sign of values. $\epsilon$ is the distortion bound. Then the obtained image is truncated to a valid range $[0,255]$.

\smallskip
\noindent{\bf BIM.} Basic Iterative Method (BIM) \cite{kurakin2016adversarial} is extended from FGSM by using iterative steps. In each iteration $t$, the adversarial image $x^{adv}$ is updated by a step $\alpha$. This process is iterated until the distortion bound $\epsilon$ is reached, as
\begin{equation}
    \begin{aligned}
         & x^{adv}_{0} = x,  \\
         & x^{adv}_{t+1} =  x^{adv}_{t} + \alpha \cdot  sign(\nabla_{x}\mc{L}(x^{adv}_{t}, y)). 
    \end{aligned}
\end{equation} 
There are two options for the selection of $\alpha$. The first is to set $\alpha = \epsilon / T$, where $T$ is the total number of iterations, and clip $x^{adv}_{t}$ to a valid range in the end. The other one is to set $\alpha = \epsilon$ and clip $x^{adv}_{t}$ at each iteration. Since BIM goes iteratively, it is more likely to find a better optimal solution than FGSM.

\smallskip
\noindent{\bf MIM.} Momentum Iterative Fast Gradient Sign Method (MIM) \cite{dong2018boosting} improves the BIM method by considering the momentum when determining the direction of descent, as
\begin{equation}
    \begin{aligned}
        & x^{adv}_{0} = x, g_0 = 0, \\
        & g_{t+1} = \mu \cdot g_t + \frac{\nabla_{x}\mc{L}(x^{adv}_{t}, y)}{|| \nabla_{x}\mc{L}(x^{adv}_{t}, y) ||_1}, \\
        & x^{adv}_{t+1} =  x^{adv}_{t} + \alpha \cdot  sign(g_{t+1}). 
    \end{aligned}
\end{equation}

\smallskip
\noindent{\bf DIM.} Diverse Inputs Iterative Fast Gradient Sign Method (DIM) \cite{xie2019improving} improves the diversity of input images by adding random transformations. Denote the random transformation operation as $\mc{T}$. The formulation of DIM is similar to MIM as
\begin{equation}
    \begin{aligned}
        & g_{t+1} = \mu \cdot g_t + \frac{\nabla_{x}\mc{L}(\mc{T}(x^{adv}_{t}), y)}{|| \nabla_{x}\mc{L}(\mc{T}(x^{adv}_{t}), y) ||_1}, \\
        & x^{adv}_{t+1} =  x^{adv}_{t} + \alpha \cdot  sign(g_{t+1}). 
    \end{aligned}
\end{equation}
At each iteration, the input image will be transformed by $\mc{T}$.

\smallskip
\noindent{\bf NIM.} Nesterov Iterative Fast Gradient Sign Method (NIM) \cite{lin2019nesterov} integrates Nesterov Accelerated Gradient (NAG) into gradient-based iterative attacks. Compared to MI-FGSM, NI-FGSM makes a step forward in the accumulated gradient direction before each iteration and then proceeds with the update, as
\begin{equation}
    \begin{aligned}
        & x^{nes}_{t} = x^{adv}_{t} + \alpha \cdot \mu \cdot g_t, \\
        & g_{t+1} = \mu \cdot g_t + \frac{\nabla_{x}\mc{L}(x^{nes}_{t}, y)}{|| \nabla_{x}\mc{L}(x^{nes}_{t}, y) ||_1}, \\
        & x^{adv}_{t+1} =  x^{adv}_{t} + \alpha \cdot  sign(g_{t+1}). 
    \end{aligned}
\end{equation}

The existing studies mainly focus on attacking classifiers but pay less attention to the vulnerability of face detectors. 



\subsection{DeepFake Defense}
\label{sec:related-works}

DeepFake detection is the classic passive defense solution to combat DeepFakes. Since the DeepFake is generated by generative models, the early forensics methods utilized traditional clues (\eg, PRNU \cite{chierchia2014bayesian}, lighting and shadow inconsistency \cite{farid2009photo}) are not effective. As such, many dedicated DeepFake detection methods have been proposed recently, \eg, \cite{li2018ictu,yang2019exposing,li2020face,han2023fcd,miao2023f,wang2023dynamic}. These methods can fall into many categories, in terms of the way they seek the forgery traces, ranging from using the biological signals \cite{li2018ictu,yang2019exposing,guo2022robust}, color signals \cite{mccloskey2018detecting,matern2019exploiting,li2020identification}, generation artifacts signals \cite{li2019exposing,li2020face} to the meticulously designed architectures and training schemes~\cite{lsda_2024_CVPR,diff-id,fei2022learning,chen2022selfsupervised,ba2024exposing, cui2024forensics} Most of these methods are based on Convolutional Neural Networks (CNNs) \cite{simonyan2014very,he2016deep} and Vision Transformers (ViT) \cite{liu2021swin,han2022survey}. They can achieve favorable, even perfect, performance on public datasets. However, they are used to lagging behind the generation of DeepFakes, unable to cut off the broadcasting of DeepFakes at the first time.

Different from DeepFake detection methods, the proactive DeepFake defense methods aim to obstruct the generation of DeepFakes. To achieve this goal, adversarial perturbations are usually utilized to malfunction the synthesis process of generative models, such as attacking the latent representations of VAEs \cite{kos2018adversarial}, attacking the image-to-image translation models in autonomous driving \cite{wang2020deceiving}, or targeting the generative models including GANs to disrupt the visual quality of synthesized faces \cite{guan2024adversarial,zhu2023information,qu2024df,huang2022cmua}. These methods focus on the generative model itself and design suitable adversarial perturbations to disrupt its original behavior. In this paper, we shift the attention from directly disrupting the DeepFake model to the data preparation step, disrupting the face detection to corrupt extracted faces. 

\section{FacePoison: Disrupting Face Detection}
\label{sec:poison}

\subsection{Threat Model}
\label{sec:method}

\subsubsection{Defender's Capacity}
The users are allowed to process their photos before uploading them online. Once the photos have been uploaded, they can hardly control where their photos will go, and whether the attackers have collected their photos to train DeepFake models. The users can also hardly intervene or access the training and testing process of the DeepFake model, \eg, the training configuration, and even the model architectures. Once the DeepFake model finishes training, the users can not intervene in the synthesis (inference) process, as this model takes the target face (not from this user) as input and outputs the victim's face (from this user). Thus, the users can not protect or are not responsible for the protection of the photos of other individuals. The described defender's capacity is the minimal requirement, being able to be applied in real-world scenarios: \textit{Users protect themselves from being the victim of DeepFakes by only manipulating their photos before uploading them online.}

\subsubsection{Defender's Goals}
By manipulating the user's photos, the faces can not be extracted by DNN-based face detectors, which contaminates the input data of the DeepFake model, subsequently obstructing their regular training or testing. Moreover, this manipulation should be imperceptible as much as possible, avoiding disturbing the original semantic content of photos.

\subsection{Problem Formulation} 
Denote $\mc{D}_{x} = \{x_i\}^N_{i=1}$ as the set of users' photos collected by attackers. Denote $\mc{F}$ as the face detector and $d_i = \mc{F}(x_i)$ denotes the detected face given image $x_i$. For simplicity, assume that only one face exists in an image. Denote $\mc{D}_d = \{d_i\}^N_{i=1}$ as the set of extracted faces using face detector $\mc{F}$ and $\mc{D}_g = \{g_i\}^N_{i=1}$ as the set of ground truth faces. Note that $\mc{D}_d$ is used to train the DeepFake model or the input faces for DeepFake faces generation. Our goal is to disrupt face detection and contaminate $\mc{D}_d$ with incorrect faces, by manipulating the images $x_i$ in $\mc{D}_{x}$ with minimal distortion, \ie, enlarging the error between $d_i$ and $g_i$. If attackers attempt to train a DeepFake model using the corrupted $\mc{D}_d$, the training process will malfunction. Similarly, if attackers use a pre-trained DeepFake model for face synthesis, the contaminated $\mc{D}_d$ will also impair the inference process.

Denote $\theta$ as the parameters of the face detector $\mc{F}$. Let $x$ be the clean image containing the victim's faces. We omit the subscript $i$ for simplicity. Disrupting face detection is to find the adversarial image $x^{adv}$ that can fool $\mc{F}$, while it is visually similar to the clean image $x$. Let $\mc{L}$ be the objective function to disrupt the face detector. This problem can be formulated as 
\begin{equation}
    \mathop{\arg\min}_{x^{adv}} \ \ \mc{L}(x^{adv}, x; \theta), \; \; s.t. \; ||x^{adv} - x||_{\infty} \leq \epsilon,
\end{equation}
where $\epsilon$ is the bound of distortion. Minimizing this equation can find an adversarial image to fool the face detector.

\begin{figure}[t]
\centering
\includegraphics[width=\linewidth]{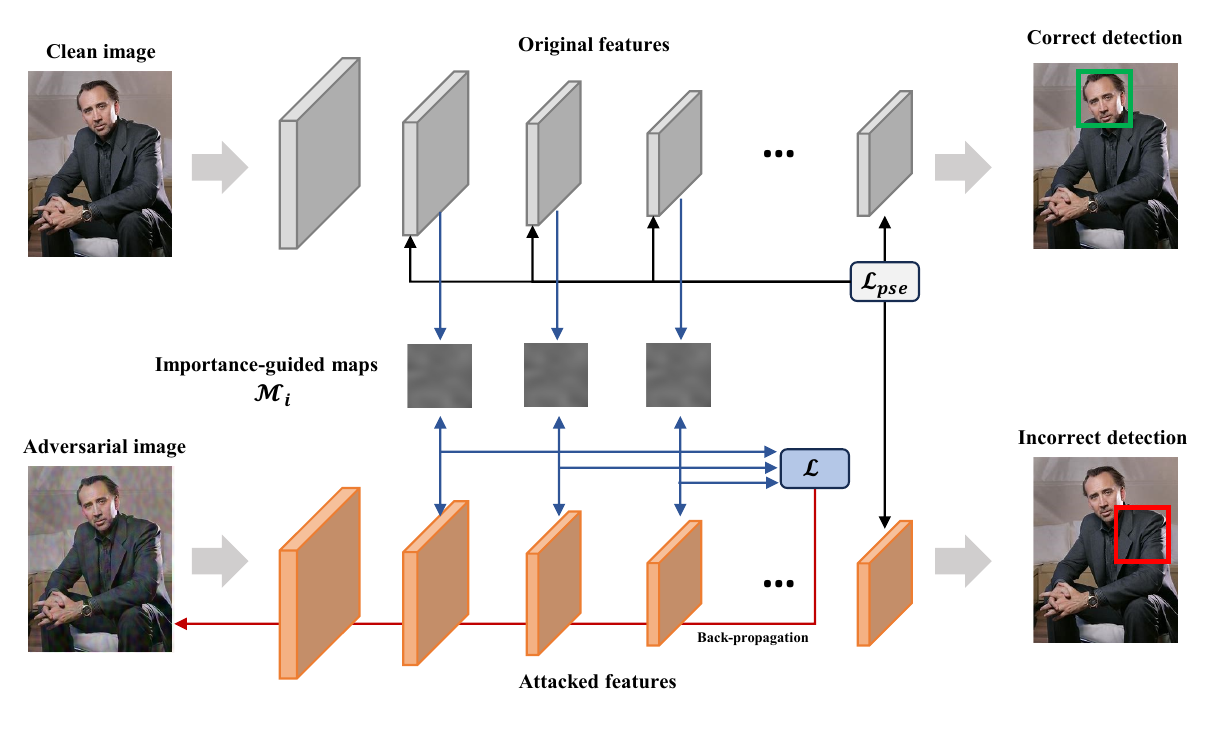}
\vspace{-1cm}
\caption{\small Overview of our method on disrupting face detection. Our method attacks multiple intermediate features with the instruction of importance-guided maps, amplifying the disturbance on key elements indicated by these maps. See text for details.}
\label{fig:methods}
\end{figure}

\subsection{Adapting Adversarial Attacks to Face Detectors}
Existing adversarial attack methods mainly focus on classifiers, which cannot be directly used to disrupt face detectors.  Therefore, we adapt the mainstream adversarial attack methods reviewed in Section~\ref{sec:adv} to fit this task. Specifically, we introduce the following modifications:

\subsubsection{Attacking Intermediate Features}
Instead of focusing on the final outputs, we target the intermediate features based on these intuitions: (1) Friendliness: Face detectors use a variety of detection heads, but their base networks are typically limited to a small number of common architectures. Targeting intermediate features ensures compatibility across different detectors. (2) Effectiveness: Intermediate features contain the important information that determines the results, thus disrupting features can naturally lead to incorrect detection. (3) Transferability: Features are more general and versatile compared to final logits, which are often overfitted to specific architectures. Thus, attacking features enhance transferability across different detectors.

Concretely, our method attacks multiple layers of the base network. Given a clean image $x$, the feature at $i$-th layer is denoted as $h_i = \mc{F}_{i}(x)$ and $\mc{H} = \{h_i \}^{K}_{i=1}$ denotes the feature set obtained from $K$ feature layers. Our method aims to mislead the face detector $\mc{F}$ by disturbing $\mc{H}$. The overview of our method is illustrated in Fig.~\ref{fig:methods}.


\subsubsection{Attack Objectives} For the conventional adversarial attacks, the objective function relates to the task, which can be easily formulated given the training samples with their corresponding annotations (\eg, labels for classification task). However, since the features are not visually understandable, how to design objectives to disturb features is important. 

In general, the elements in features have different impacts on the results. Disturbing the high-impact elements can be more effective in disrupting the results. Inspired by~\cite{wang2021feature,he2023improving}, we design an importance-guided map for each scale. The larger value in this map represents the greater importance of corresponding feature elements. Denote $\mc{H}' = \{h'_i \}^{K}_{i=1}$ as the corresponding set given the adversarial image $x^{adv}$. Our objective function can be defined as 
\begin{equation}
    \mc{L}(x^{adv}, x; \theta) = \sum_{i \in \phi} \alpha_i \cdot (\mc{M}_i \cdot h'_i),
    \label{equ:objective}
\end{equation}
where $\phi \subseteq \{1,2,...,K \}$ is a subset of all feature layer indexes, $\alpha_i$ is the weight factor for each scale, $\mc{M}_i$ is the importance-guided map for each scale. Minimizing this equation can decrease the response of important feature elements while increasing the response of unimportant feature elements, effectively disrupting the normal distribution of features. 

One straightforward way to decide whether a feature element is important is to back-propagate the gradients from the task-related objective of training the face detectors to the features $\mc{H}'$. However, the objective of training face detectors likely varies due to their different architectures and training schemes, hindering the generalization of attacks, as the type of face detectors needs to be known in advance. As such, we propose a pseudo-objective that considers feature differences to replace the task-related objective, disentangling the relationship between the attack and its original task-related objective. Specifically, we select the last feature layer $h_K$ from the clean feature set $\mc{H} = \{h_i \}^{K}_{i=1}$ as a reference. Then we calculate the distance between this clean feature layer $h_K$ and the attacked feature layer $h'_K$ to approximate the error in the task-related objective. We use the last feature layer for the pseudo-objective as the deep layer has more concentrated, high-level semantic information than the shallow layer, which can better depict the task-related objective. We use cosine similarity to measure the feature difference as 
\begin{equation}
    \mc{L}_{pse}(x^{adv}, x; \theta) = \frac{h_K \cdot {h'_K}^{\top}}{||h_K||\cdot ||{h'_K}||}.
\end{equation}
For a feature layer $h'_i$ that is going to be attacked, we can obtain the importance-guided map $\mc{M}_i$ by back-propagating the gradient of $\mc{L}_{pse}$ to the $i$-th layer as 
\begin{equation}
    \mc{M}_i = \frac{\partial \mc{L}_{pse}(x^{adv}, x; \theta)}{\partial h'_i}.
    \label{equ:m-i}
\end{equation}
Since $\mc{M}_i$ is composed of the gradients, it may have noise specific to the model architecture. To reduce these noises, we apply random masking on the input image for $m$ times and calculate the importance-guided map $\mc{M}'_i$ for each time. The random masking is to set the pixel value of the input image to zero with the probability $p$. Then we average these importance-guided maps and normalize the averaged map as the final $\mc{M}_i$.

\begin{figure}[t]
\centering
\includegraphics[width=\linewidth]{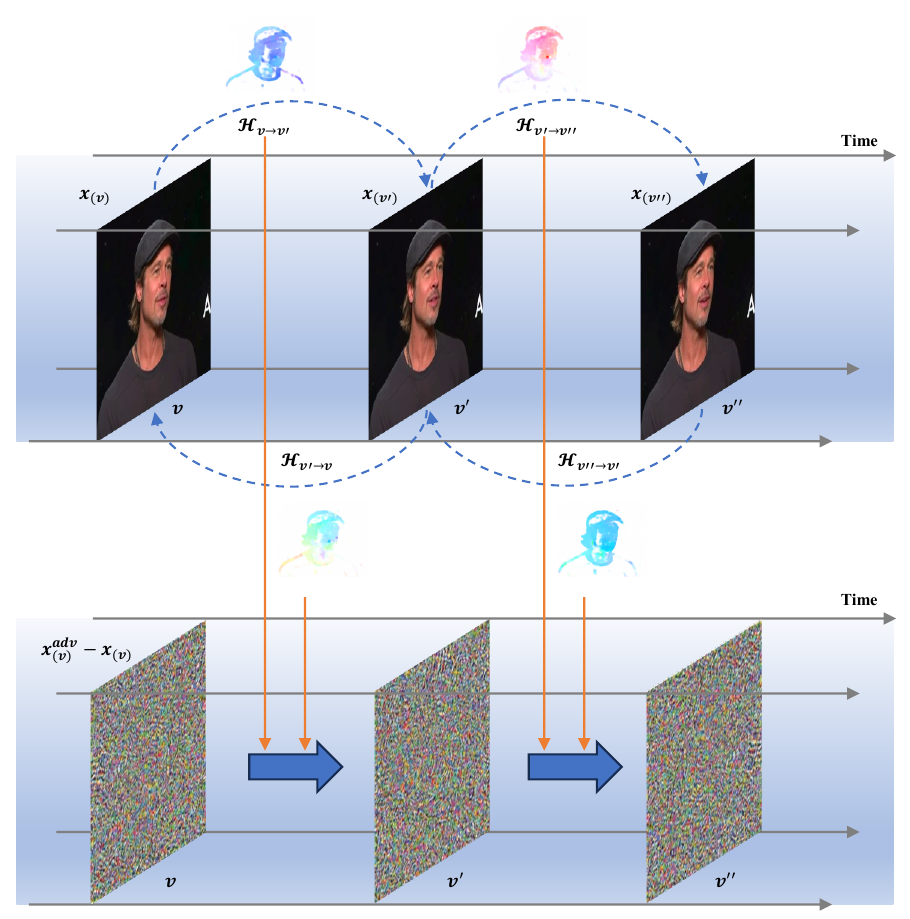}
\caption{\small Overview of the VideoFacePoison method on disrupting face detection in a video. Our method can propagate the FacePoison on a certain frame ($x_{(v)}$ in this example) to adjacent frames based on the optical flow algorithms. See text for details.}
\label{fig:videofacepoison}
\end{figure}

\section{VideoFacePoison: Propagating FacePoison across Frames in Video}
\label{sec:videofacepoison}
Based on FacePoison, we describe VideoFacePoison to protect videos by propagating adversarial perturbations across frames. While videos can be protected by directly applying perturbations to each frame, this process is computationally expensive. By leveraging the temporal correlations between adjacent frames, we are inspired to improve the efficiency of video protection.

\hypo {\em The temporal relationships across adjacent frames can also be represented on adversarial perturbations. Therefore, the adversarial perturbations created on one frame can be propagated to adjacent frames while retaining their effectiveness. }

This hypothesis stems from the observation that the adversarial perturbations are highly related to image content~\cite{chakraborty2022generalizing} and can be estimated from nearby frames using the optical flow algorithms. Thus, we utilize optical flow algorithms to estimate and propagate adversarial perturbations.

\subsubsection{Revisit of Optical Flow} Basically, optical flow indicates the displacement of objects among adjacent frames. Optical flow algorithms assume that the intensity of pixels in an object remains unchanged among adjacent frames. Denote the intensity of a pixel at location $(x, y)$ and time $t$ as $I(x,y,t)$. We reuse the notation $x,t$ here to illustrate optical flow better. Denote $\Delta x$ and $\Delta y$ are the displacements during time $\Delta t$. Under the above assumption, an equation can be written as $I(x,y,t) = I(x+\Delta x, y+\Delta y, t+\Delta t)$. Then we can obtain $0=\frac{\partial I}{\partial x} \Delta x + \frac{\partial I}{\partial y} \Delta y + \frac{\partial I}{\partial t} \Delta t$ using Taylor expansion. Many methods exist to calculate optical flow, \eg, 
Gunnar Farneback \cite{farneback2003two}, FlowNet \cite{dosovitskiy2015flownet}, etc.

\subsubsection{Method Introduction} Denote $\mc{X} = \{x_{(v)} \}_{v=0}^{V}$ as a video, and $v$ is the time index of the frame in this video. The optical flow map from the $v$-th frame to the $v'$-th frame can be denoted as $\mc{H}_{v \rightarrow v'}$. Given the FacePoison at $v$-th frame, we can estimate the FacePoison for the adjacent $v'$-th frame by mapping the perturbations based on $\mc{H}_{v \rightarrow v'}$ as 
\begin{equation}
    x^{adv}_{(v')} = \mc{P}(x^{adv}_{(v)} - x_{(v)}, \mc{H}_{v \rightarrow v'}) + x_{(v')},
    \label{eq:forward}
\end{equation}
where $\mc{P}$ is a mapping function that can map the pixels in the current frame to adjacent frames based on the optical flow in the forward direction. As the optical flow algorithms are not perfect, they can also introduce artifacts. To reduce the artifacts, we also consider the optical flow of backward direction as $\mc{H}_{v' \rightarrow v}$, and can obtain the prediction at $v'$-th frame by reversing the $\mc{H}_{v' \rightarrow v}$ as 
\begin{equation}
    x^{adv'}_{(v')} = \mc{P}(x^{adv}_{(v)} - x_{(v)}, -\mc{H}_{v' \rightarrow v}) + x_{(v')}.
\end{equation}
Then we average these two predictions as the FacePoison for the $v'$-th frame. This propagation will stop after a certain time in case the faces have an apparent displacement, and then we recalculate the FacePoison and propagate it again. This procedure will be repeated until all frames are processed. The overview of our VideoFacePoison method is shown in Fig.~\ref{fig:videofacepoison}.

\section{Experiments}
\label{sec:exp}

\subsection{Experimental Settings}
\label{sec:settings}
\subsubsection{Datasets}
We employ WIDER~\cite{yang2016wider}, a widely used face detection dataset, to demonstrate the efficacy of our methods. This dataset contains plenty of faces with location annotations. Moreover, to evaluate the efficacy in obstructing DeepFake generation, we conduct validations of our method using four standard DeepFake datasets, including FaceForensics++ (FF++)~\cite{rossler2019faceforensics++}, Celeb-DF~\cite{li2020celeb}, DeepFake Detection Challenge (DFDC) dataset\cite{dolhansky2020deepfake} and preview version of DFDC (DFDCP) dataset\cite{dolhansky2019dee}.

\subsubsection{Face Detectors}
We consider five mainstream DNN-based face detectors in experiments, which are  \textbf{RetinaFace} \cite{deng2020retinaface}, \textbf{YOLO5Face} \cite{qi2022yolo5face}, \textbf{PyramidBox} \cite{tang2018pyramidbox}, \textbf{S3FD} \cite{zhang2017s3fd}, and \textbf{DSFD} \cite{li2019dsfd}, respectively. RetinaFace is built upon MobileNet-0.25 \cite{howard2017mobilenets}. YOLO5Face is redesigned from the YOLOV5 object detector \cite{YOLOv5}, which is based on ShuffleNetv2 \cite{zhang2018shufflenet}. PyramidBox, S3FD, and DSFD utilize VGG16 as their basenetworks. All of these face detectors are trained under their default settings.

\subsubsection{Adapted Adversarial Attack Methods}
We adapt the adversarial attack methods {BIM} \cite{kurakin2016adversarial}, {DIM} \cite{xie2019improving}, {MIM} \cite{dong2018boosting}, and NIM~\cite{lin2019nesterov} into this task, using the proposed modifications. We denote these adapted methods as \textbf{Ada-BIM}, \textbf{Ada-DIM}, \textbf{Ada-MIM}, and \textbf{Ada-NIM}, respectively. Moreover, inspired by~\cite{1672377,wang2021feature}, we incorporate operations of spectrum transformation and gradient averaging operations into DIM to improve the robustness. We term this incorporation as \textbf{Ada-DIM++}. Note that spectrum transformation increases data diversity by randomly modifying the frequency components of the input, while gradient averaging helps stabilize the attack direction.  

\subsubsection{DeepFake Models}
To fully demonstrate the effectiveness of our method, we evaluate our method on \textbf{eleven} modern DeepFake models, including four representative academic methods: \textbf{SimSwap} (ACM MM'20)\cite{chen2020simswap}, \textbf{InfoSwap} (CVPR'21) \cite{Gao_2021_CVPR}, \textbf{MobileFaceSwap} (AAAI'22) \cite{xu2022mobilefaceswap}, \textbf{BlendFace} (ICCV'23) \cite{shiohara2023blendface}, five variants from the open-source tool \textit{FaceSwap} \cite{faceswap} (namely \textbf{Origin}, \textbf{DFaker}, \textbf{IAE}, \textbf{LightWeight}, and \textbf{DFLH}) and two additional models from \cite{li2020celeb} (namely \textbf{CDFv1} and \textbf{CDFv2}). 
SimSwap\footnote{\url{https://github.com/neuralchen/SimSwap}} adopts an auto-encoder structure with an Identity Injection Module to preserve identity features during synthesis. InfoSwap\footnote{\url{https://github.com/GGGHSL/InfoSwap-master}} introduces an Informative Identity Bottleneck and Adaptive Information Integrator to decouple and integrate identity information. MobileFaceSwap\footnote{\url{https://github.com/Seanseattle/MobileFaceSwap}} is a lightweight model that leverages knowledge distillation to achieve high-quality face swapping with reduced computational cost. BlendFace\footnote{\url{https://github.com/mapooon/BlendFace}} separates identity and attribute features using a dedicated face recognition module and identity encoder for better face synthesis. The FaceSwap variants (Origin, DFaker, IAE, LightWeight, DFLH) share a typical encoder-decoder framework, with variations in synthesis resolution, encoder/decoder architectures, and training strategies. Detailed configurations are available in the official repository. Similarly, {CDFv1} and {CDFv2} are designed following FaceSwap with improvement on synthesis size and spatial-temporal alignment. 
We also note that \textit{DeepFaceLab}~\cite{DeepFaceLab} is another open-source tool for face swapping. Since this repository shares a similar generation pipeline to FaceSwap, we use FaceSwap for demonstration for simplicity.

\subsubsection{Evaluation Metrics} Since our goal is to obstruct the DeepFake generation by disrupting face detection, we evaluate our method from two perspectives: whether the face detection is disrupted and whether the DeepFake generation is obstructed. We use F1-score to evaluate the attacking performance of our method, which is calculated as $$\text{F1-score} = 2 \cdot \frac{\text{Recall} \cdot \text{Precision}}{\text{Recall} + \text{Precision}}.$$ F1-score considers both the precision and recall, which is more comprehensive than the Average Precision (AP) metric (See Section \ref{sec:discussion} for more details). The lower score of these metrics means the faces are not well detected. To evaluate the performance of DeepFake obstruction, we employ the SSIM score to measure the visual quality of the generated DeepFake. The larger score indicates that the quality of DeepFake faces is better.

\subsubsection{Implementation Details} The experiments are conducted on Ubuntu 22.04 with one Nvidia 3060 GPU. The number of feature layers is set to $K=3$. We target specific intermediate feature layers for each face detector architecture to conduct the attack. For PyramidBox, we target the feature outputs following the ReLU activations after conv3\_3 and conv4\_3, and the convolutional output of conv5\_3. These correspond to the 15th, 22nd, and 28th feature layers in the VGG16 backbone, respectively. For S3FD and DSFD, we target three feature layers: conv2\_2 (7th), pool3 (16th), and a high-level context convolutional layer (31st). For YOLO5Face, we target the feature outputs after the final SiLU activations in the first three ShuffleNetV2 blocks of the backbone. For RetinaFace, we target the feature outputs after the final LeakyReLU activations in the three stages (stage1, stage2, and stage3) of its MobileNet-0.25 backbone. The parameter settings for FacePoison are as follows: $\alpha_1 = 0.2, \alpha_2 = 0.3, \alpha_3 = 0.5, p=0.9, m=30$. In Ada-DIM++, we introduce spectrum transformation and gradient averaging. The standard deviation is set to $\sigma=16$, the fluctuation range to $\mu=0.5$, the number of spectrum transformations per iteration to $n=10$, and the gradient averaging window to $o=4$. The maximum iteration number is $10$, and the bound of distortion is set to $\epsilon = 8$. Note that the bound of distortion in attacking general image classification is usually set to $16$, see \cite{kurakin2016adversarial,dong2018boosting,xie2019improving}. Given the higher sensitivity of face images, we restrict the bound to $8$ to maintain imperceptibility. For VideoFacePoison, we utilize Gunnar Farneback \cite{farneback2003two} to calculate optical flow.

\subsection{Results of Disrupting Face Detection}


\begin{table}[!t]
    \centering
    \caption{\small F1-score ($\%$) of different methods on WIDER dataset. RF, YF, and PB denote RetinaFace, YOLO5Face, and PyramidBox, respectively.}
    \vspace{-0.3cm}
    \label{tab:attack_fd1}
    \begin{tabular}{c|ccccc}
    \hline
    Methods  & RF & YF & PB & S3FD & DSFD  \\
    \hline 
    None    & 94.0 & 98.4 & 98.7 & 98.0 & 98.6 \\
    Random  & 93.8 & 98.3 & 98.7 & 98.2 & 98.3 \\
    Ada-BIM     & 1.4 & 0.0 & 1.7 & 1.6 & 0.3 \\ 
    Ada-DIM     & 9.7 & 0.8 & 18.8 & 53.0 & 41.2 \\ 
    Ada-MIM     & 1.5 & 0.0 & 3.5 & 13.3 & 3.8 \\
    Ada-NIM     & 1.6 & 0.0 & 3.5 & 13.6 & 3.6 \\
    Ada-DIM++   & 0.6 & 0.3 & 0.4 & 3.9 & 5.0 \\ 
    \hline
    \end{tabular}
    
\end{table}

\subsubsection{Results}
\begin{figure}[t]
    \centering
    \includegraphics[width=\linewidth]{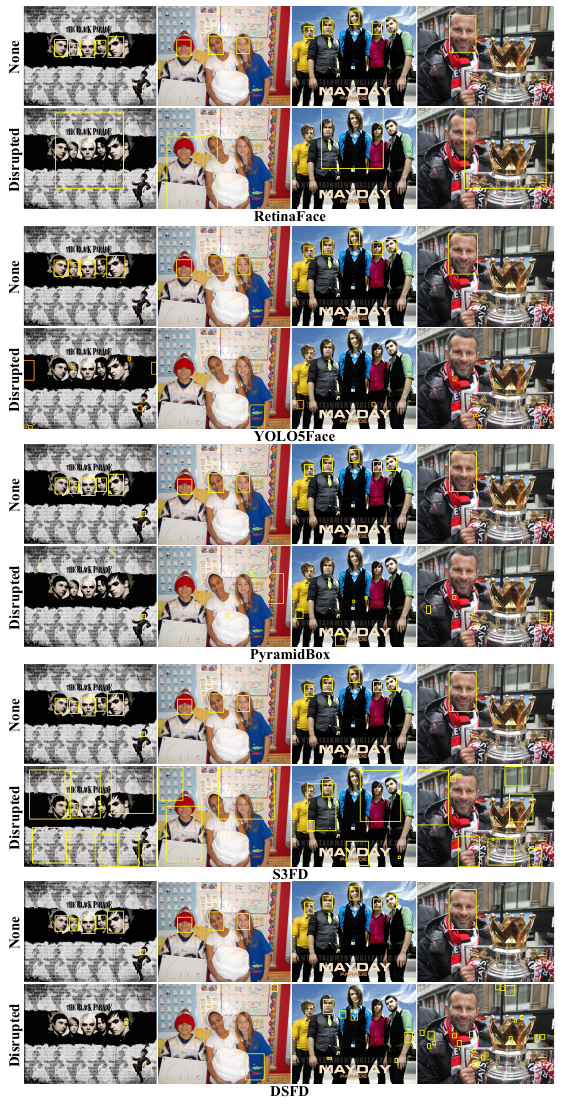}
    \vspace{-0.9cm}
    \caption{\small Visual examples of using different adaptations to disrupt face detection. For each face detector, the top row corresponds to the original results and the bottom row is the results using our method.}
    \label{fig:attack_demos}
\end{figure}

Table \ref{tab:attack_fd1} shows the F1-score ($\%$) of different methods attacking face detectors on the WIDER dataset. RF, YF, and PB denote RetinaFace, YOLO5Face, and PyramidBox, respectively. None denotes that no noises are added to images. By adding random noises, the performance of these methods merely drops, which indicates their robustness against random noises. It can be seen that the adapted methods Ada-BIM, Ada-DIM, Ada-MIM, Ada-NIM, and Ada-DIM++ can effectively disrupt face detection results. Several visual examples are illustrated in Fig.~\ref{fig:attack_demos}.

\subsubsection{Ablation Study} Using Ada-DIM++ as an example, we study the effect of attacking multiple layers on the WIDER dataset. Specifically, we conduct experiments by attacking various numbers of layers. The results are shown in Table~\ref{tab:abs}. Here, L1, L2, and L3 denote the first, second, and third layers attacked by our method. The top three rows show the performance when only a single layer is attacked, while the bottom row corresponds to our method of attacking multiple layers. We can observe that disturbing all three layers achieves the best performance compared to single layer attacks, demonstrating the efficacy of targeting multiple layers.


\begin{table}[t]
    \centering
    \caption{\small Effect of attacking multiple layers. }
    \vspace{-0.3cm}
    \label{tab:abs}
    \begin{tabular}{c|cccccc}
    \hline
      & RF & YF & PB & S3FD & DSFD  \\
    \hline 
    L1 & 61.9 & 57.2 & 87.2 & 98.3 & 98.5 \\ 
    L2 & 0.7 & 0.7 & 15.7 & 67.2 & 84.8 \\ 
    L3 & 0.7 & 0.8 & 0.5 & 3.8 & 9.1 \\ 
    L1,L2,L3 & 0.6 & 0.3 & 0.4 & 3.9 & 5.0 \\ 
    \hline
    \end{tabular}

\end{table}

\begin{table}[t]
    \centering
    \caption{\small Transferability of our method on disrupting face detection. ``S'' and ``T'' denote source and target face detectors, respectively.}
    \vspace{-0.3cm}
    \label{tab:trans}
    
    \begin{tabular}{c|ccccc}
    \hline
    S$\downarrow$, T$\rightarrow$ & RF & YF & PB & S3FD & DSFD  \\
    \hline
    RF      & 0.6 & 89.8 & 97.6 & 96.4 & 98.4 \\ 
    YF   & 84.2 & 0.3 & 97.1 & 97.0 & 98.3 \\ 
    PB & 76.8 & 72.2 & 0.4 & 8.3  & 2.9 \\ 
    S3FD       & 66.9 & 77.3 & 20.9 & 3.9 & 16.5 \\ 
    DSFD       & 70.6 & 70.3 & 11.3 & 9.4  & 5.0 \\ 
    \hline 
    
    \hline
    \end{tabular}
\end{table}

\begin{figure}[t]
    \centering

    \includegraphics[width=0.45\linewidth]{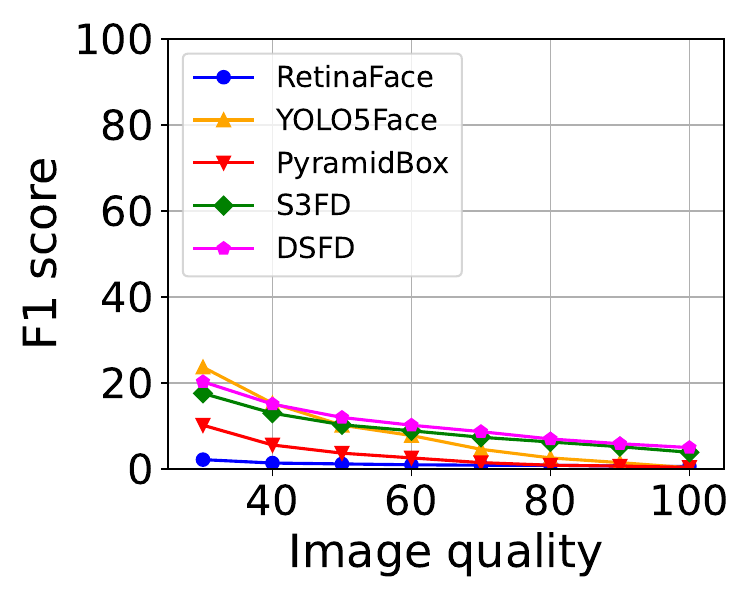}
    \includegraphics[width=0.45\linewidth]{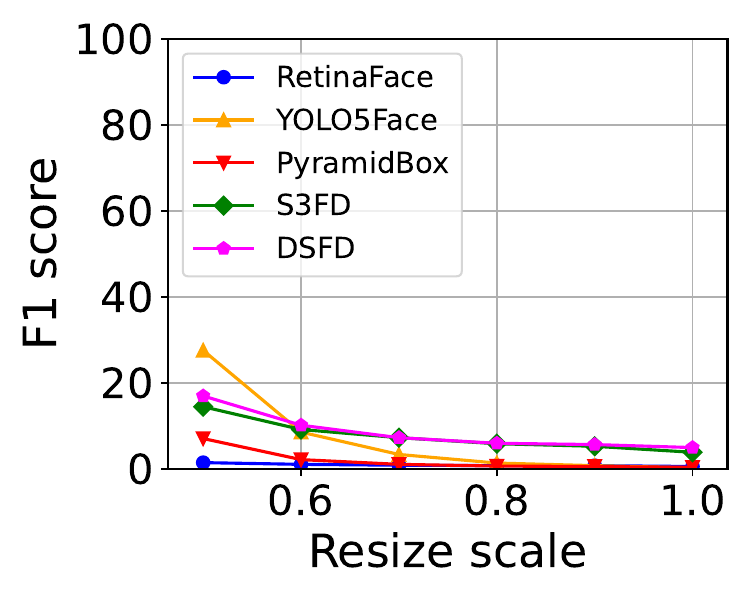} \\
    \includegraphics[width=0.45\linewidth]{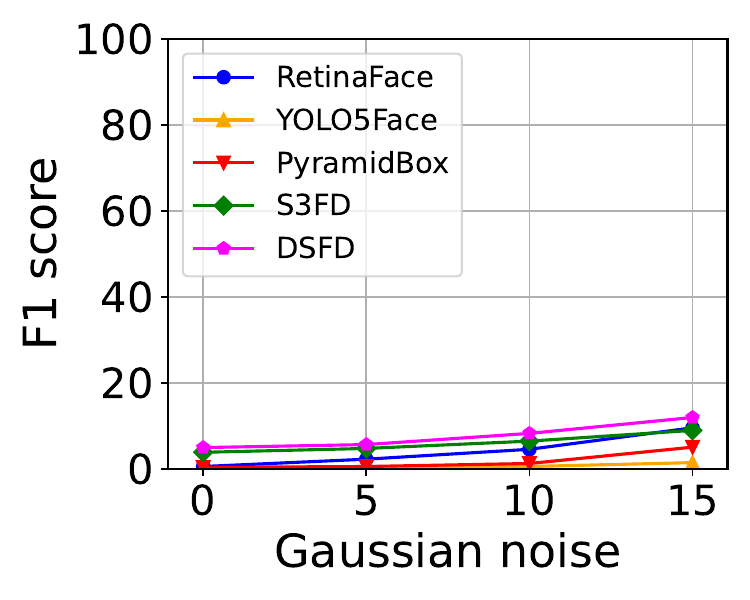}
    \includegraphics[width=0.45\linewidth]{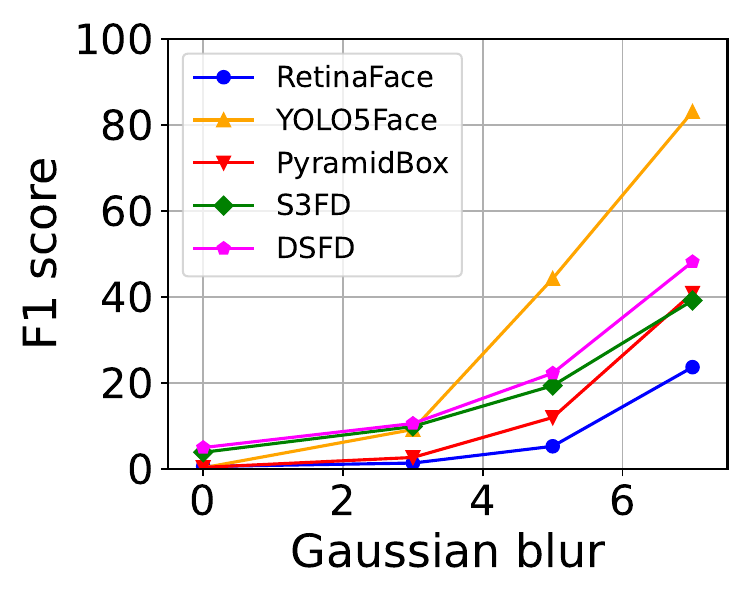}
    \vspace{-0.4cm}
    \caption{\small Performance of different face detectors under different image quality, resize scale, Gaussian noise, and Gaussian blurring.}
    \label{fig:robustness}
\end{figure}

\begin{table}[t]
    \centering
    \caption{\small  The performance of different face detectors under various countermeasures.}
    \vspace{-0.3cm}
    \label{tab:robustness_new}
    \begin{tabular}{c|ccccc}
    \hline
     & RF & YF & PB & S3FD & DSFD  \\
    \hline 

    None     & 94.0  & 98.4  & 98.7  & 98.0  & 98.6 \\ 
    Ada-DIM++ & 0.6 & 0.3 & 0.4  & 3.9 & 5.0 \\
    \hline
    NRP & 54.5 & 72.5 & 86.2 & 78.5 & 81.7 \\
    Facebook &  0.8 & 1.1 & 0.6 & 5.0 & 5.9 \\
    Twitter & 0.8 & 0.7 & 0.6 & 4.6 & 5.4 \\ 
    \hline
    \end{tabular}
\end{table}


\subsubsection{Transferability}
We further explore the transferability of our method, using Ada-DIM++ as an example. Transferability refers to the ability to attack one face detector (target) using adversarial perturbations generated from another face detector (source). The results are shown in Table \ref{tab:trans}, where ``S'' and ``T'' denote source and target face detectors, respectively. Our method demonstrates transferability in most cases. For example, using DSFD to attack RetinaFace can notably reduce the performance from $94.0\%$ to $70.6\%$ while attacking YOLO5Face degrades its performance from  $98.4\%$ to $70.3\%$. The transferability is particularly pronounced among PyramidBox, S3FD, and DSFD, greatly decreasing the performance below $21\%$. However, when RetinaFace or YOLO5Face is used as the source to attack the other three detectors, the transferability is ineffective. We hypothesize that this is due to the lightweight nature of these source detectors compared to the target detectors, which hinders knowledge transfer, thus limiting the effectiveness of adversarial perturbations.

\subsubsection{Robustness}\label{sec:robustness}
Since the social platform usually processes the uploaded images, we study the robustness of our method in this part, including the resistance to image compression, resizing, Gaussian noise, and Gaussian blurring.
{1) Image compression}: We use OpenCV to control image quality, varying the compression factor from $[30, 100]$, where a larger value indicates higher quality. Fig. \ref{fig:robustness} (top left) shows the results of our method confronting different image compression operations. We can see that the F1-score of these face detectors is still around $10\%$ at quality $50$, indicating resilience to image compression.
{2) Resizing}: To counter adversarial perturbations, transformation-based defenses such as random resizing and padding~\cite{xie2017mitigating} are commonly employed. We randomly resize input images with a ratio in {$[0.5, 1]$}. As shown in Fig.~\ref{fig:robustness} ({top} right), the F1-score decreases only slightly as the resizing ratio decreases, demonstrating a degree of robustness. 
3) Gaussian noise: We add random Gaussian noise with standard deviations of 5, 10, and 15 to simulate varying levels of distortion. Fig. \ref{fig:robustness} (bottom left) shows that our method remains effective even at higher noise levels. 4) Gaussian blur: Robustness to Gaussian blur is evaluated using different kernel sizes.  As shown in Fig. \ref{fig:robustness} (bottom right), our method maintains good performance with kernel sizes of 3 and 5, further highlighting its robustness.

To further study the robustness, we apply countermeasures such as adaptive attacks and platform-level defenses on our method. Specifically, we employ a representative adaptive attack \textit{NRP}, an adversarially trained purifier proposed in~\cite{naseer2020self} to eliminate our perturbations. For platform-level defenses, we select Facebook and Twitter as our target social media platforms. Due to upload limitations on these platforms, we follow the offline processing strategy described in~\cite{qu2024df}. 
The detailed experimental results are shown in Table~\ref{tab:robustness_new}. None indicates original detection performance without any perturbations. Ada-DIM++ is our method for disrupting face detectors. The results show that NRP partially mitigates the effect of our method, recovering some functionality of face detectors. Nevertheless, our method still has an effect under this defense. In contrast, the processing in social platforms has a negligible effect on our method, demonstrating its robustness in real-world usage scenarios. 

\begin{table}[!t]
    \centering
    \caption{\small F1-score ($\%$) of VideoFacePoison and its variants. FP-all denotes applying the FacePoison on every frame. FP-fixed denotes applying the FacePoison of the first frame to all other frames. FP-forward represents only considering the optical flow in the forward direction.}
    \vspace{-0.3cm}
    \label{tab:videofacepoison}

    \begin{tabular}{c|ccccc|c}
    \hline
     & RF & YF & PB & S3FD & DSFD & Avg. \\
    \hline 
    FP-all     & 0.0  & 0.4  & 1.0  & 2.6  & 23.4 & 5.48 \\ 
    FP-fixed   & 47.4 & 28.1 & 39.1 & 33.1 & 51.8 & 39.9 \\
    FP-forward & 46.7 & 55.8 & 22.9 & 20.2 & 44.2 & 38.0 \\
    CADVG~\cite{liu2023coherent} & 56.1 & 69.0 & 0.7 & 63.1 & 0.0 & 37.8 \\
    \hline
    VideoFacePoison        & 33.6 & 59.0 & 19.6 & 15.0 & 45.5 & 34.5 \\ 
    \hline
    \end{tabular}
\end{table}

\subsubsection{Results of VideoFacePoison}
Since the WIDER dataset is image-based, we use the videos in the Celeb-DF dataset to validate our method. Specifically, we select three identities (id0, id1, id2), each with 10 videos. Since these videos do not have face annotations, we use the results of face detectors as the ground truth. For each video, we first calculate the FacePoison on the first frame and then propagate this perturbation across frames. As the subject in a video moves very smoothly with minimal changes between adjacent frames, we evaluate the performance of our method with an interval of 5 frames. After a certain propagation length, we recalculate FacePoison on a new frame and repeat the process until all frames are processed. 

Table~\ref{tab:videofacepoison} shows the performance of VideoFacePoison. FP-all denotes that we apply FacePoison on every frame. FP-fixed means we apply the FacePoison of the first frame to all other frames. FP-forward represents that we only consider the optical flow from the current frame to the next frame using Eq.~\eqref{eq:forward}. It is important to note that CADVG~\cite{liu2023coherent} also uses optical flow in adversarial perturbation generation, but it is originally designed to attack DeepFake detectors, not face detectors. Unlike our method, it requires creating adversarial perturbations for every video frame, using optical flow as a restriction loss to smooth perturbations across adjacent frames. For a comprehensive comparison, we adapt this method to our task. Since the authors have not released the source code, we implement the algorithm rigorously following their description in the paper.
It can be seen that using FP-all can achieve the best performance compared to others if the time cost is not considered. Since the subject in the video does not have dramatic movement, FP-fixed also has a decent effect on detection disruption. In comparison, FP-forward propagates the adversarial perturbations to adjacent frames instead of using the fixed pattern, it surpasses the FP-fixed on average. Our method, which incorporates both forward and backward optical flow for propagation, largely outperforms FP-forward on average. Compared to CADVG, our method can achieve better performance, even though it involves additional adversarial optimization on every frame. Note that on YOLO5Face, the optical flow propagation is less effective. This is likely due to the base architecture, YOLOv5, which relies on complex feature fusion and multi-scale processing when predicting detection boxes, which requires more precise adversarial perturbations. Consequently, propagated perturbations may lose effectiveness. The similar trend is also observed in CADVG. Despite this, the average results demonstrate that our optical flow-based propagation strategy remains effective for disrupting face detection, aligning with our hypothesis that adversarial perturbations strongly correlate with the semantic content of the images. 

Moreover, we evaluate the computational efficiency of the proposed VideoFacePoison method. The results are shown in Table~\ref{tab:videofacepoison_time}. It can be seen that the improved VideoFacePoison method is significantly more efficient than FP-all, due to the newly proposed optical flow strategy, which enables effective propagation of perturbations across video frames. Additionally, when compared to CADVG, our method still achieves faster runtime, demonstrating superior computational efficiency.

\begin{table}[t]
    \centering
    \caption{\small  The efficiency (seconds per frame) of VideoFacePoison and other methods across different face detectors.}
    \vspace{-0.3cm}
    \label{tab:videofacepoison_time}
    \begin{tabular}{c|ccccc|c}
    \hline
     & RF & YF & PB & S3FD & DSFD & Avg. \\
    \hline 

    FP-all     & 1.18  & 1.60  & 13.13  & 10.53  & 10.92 & 7.47 \\ 
    CADVG & 0.36 & 0.39 & 2.11 & 1.32 & 2.07 & 1.25 \\
    VideoFacePoison        & 0.21 & 0.19 & 0.36 & 0.28 & 0.29 & 0.27 \\ 
    \hline
    \end{tabular}
\end{table}

\begin{table*}[!ht]
    \centering
    \caption{\small Obstructing inference: SSIM score ($\%$) of synthesized faces by obstructing the inference phase of DeepFake models using different face detectors on different datasets. AO: the source identity is polluted while the target identity is original. OA: the source identity is original while the target identity is polluted. }
    \vspace{-0.3cm}
    \label{tab:ssim_inference}
    \begin{tabular}{c|c|c|cc|cc|cc|cc|cc}
    \hline
    \multirow{2}{*}{Dataset} 
    & \multirow{2}{*}{DF model}
    & \multirow{2}{*}{Method}
    & \multicolumn{2}{c|}{RetinaFace}
    & \multicolumn{2}{c|}{YOLO5Face}
    & \multicolumn{2}{c|}{PyramidBox}
    & \multicolumn{2}{c|}{S3FD}
    & \multicolumn{2}{c}{DSFD} \\
    \cline{4-13}
    & & & AO & OA & AO & OA & AO & OA & AO & OA & AO & OA \\
    \hline
    \multirow{8}{*}{FF++}
     & \multirow{2}{*}{SimSwap}
     & Random   & 99.8 & 88.4 & 99.8 & 88.5 & 99.8 & 88.8 & 99.8 & 87.7 & 99.8 & 88.2 \\
     & & Ours     & 71.1 & 10.0 & 39.1 & 24.7 & 83.7 & 24.8 & 88.7 & 15.6 & 83.9 & 7.5  \\
     \cline{2-13}
     & \multirow{2}{*}{InfoSwap}
     & Random   & 98.3 & 88.7 & 98.3 & 88.6 & 98.4 & 89.2 & 98.4 & 88.3 & 98.4 & 88.6  \\
     & & Ours     & 66.0 & 18.8 & 36.8 & 33.0 & 78.2 & 39.4 & 84.0 & 27.3 & 79.4 & 12.9  \\
     \cline{2-13}
     & \multirow{2}{*}{MobileFaceSwap}
     & Random   & 99.9 & 88.9 & 99.9 & 89.1 & 99.9 & 89.3 & 99.9 & 88.3 & 99.9 & 88.8  \\
     & & Ours     & 72.8 & 11.2 & 39.8 & 25.1 & 85.3 & 26.7 & 90.6 & 16.9 & 84.8 & 8.0  \\
     \cline{2-13}
     & \multirow{2}{*}{BlendSwap}
     & Random   & 99.7 & 87.2 & 99.7 & 87.4 & 99.6 & 87.7 & 99.7 & 86.5 & 99.7 & 87.0  \\
     & & Ours     & 75.1 & 12.4 & 40.6 & 25.2 & 88.6 & 27.1 & 94.2 & 18.8 & 85.5 & 8.8  \\
     \cline{1-13}
     \multirow{8}{*}{CDF-v2}
     & \multirow{2}{*}{SimSwap}
     & Random   & 99.8 & 88.9 & 99.7 & 88.3 & 99.6 & 88.7 & 99.7 & 88.8 & 99.6 & 88.6  \\
     & & Ours     & 74.0 & 14.1 & 47.5 & 18.7 & 82.6 & 28.3 & 88.9 & 27.6 & 66.8 & 33.4  \\
     \cline{2-13}
     & \multirow{2}{*}{InfoSwap}
     & Random   & 98.6 & 92.2 & 98.5 & 91.8 & 98.4 & 92.0 & 98.5 & 92.2 & 98.4 & 92.1  \\
     & & Ours     & 70.3 & 28.6 & 44.9 & 28.5 & 77.9 & 43.6 & 84.1 & 43.6 & 63.2 & 48.5  \\
     \cline{2-13}
     & \multirow{2}{*}{MobileFaceSwap}
     & Random   & 99.8 & 89.7 & 99.9 & 89.0 & 99.8 & 89.4 & 99.8 & 89.5 & 99.8 & 89.4  \\
     & & Ours     & 76.4 & 15.3 & 49.2 & 19.2 & 85.2 & 29.5 & 91.7 & 29.0 & 68.8 & 35.0  \\
     \cline{2-13}
     & \multirow{2}{*}{BlendSwap}
     & Random   & 99.7 & 88.4 & 99.7 & 87.8 & 99.6 & 88.2 & 99.6 & 88.3 & 99.6 & 88.1  \\
     & & Ours     & 78.7 & 18.4 & 50.2 & 20.1 & 87.1 & 31.0 & 93.8 & 30.3 & 70.0 & 36.1  \\
     \cline{1-13}
     \multirow{8}{*}{DFDC}
     & \multirow{2}{*}{SimSwap}
     & Random   & 99.6 & 90.7 & 99.6 & 90.0 & 99.7 & 90.2 & 99.6 & 90.4 & 99.6 & 90.3  \\
     & & Ours    & 26.8 & 18.8 & 24.6 & 28.9 & 35.2 & 35.9 & 92.0 & 23.0 & 77.2 & 34.7  \\
     \cline{2-13}
     & \multirow{2}{*}{InfoSwap}
     & Random   & 98.5 & 94.0 & 98.5 & 93.5 & 98.6 & 93.7 & 98.5 & 93.8 & 98.4 & 93.7  \\
     & & Ours     & 26.1 & 30.0 & 23.9 & 39.0 & 34.5 & 49.5 & 89.0 & 35.3 & 74.4 & 48.3  \\
     \cline{2-13}
     & \multirow{2}{*}{MobileFaceSwap}
     & Random   & 99.8 & 91.4 & 99.7 & 90.7 & 99.8 & 90.8 & 99.7 & 91.1 & 99.7 & 90.9  \\
     & & Ours     & 27.2 & 19.8 & 25.1 & 29.8 & 35.9 & 36.3 & 93.6 & 23.4 & 78.7 & 35.6  \\
     \cline{2-13}
     & \multirow{2}{*}{BlendSwap}
     & Random   & 99.7 & 90.0 & 99.7 & 89.3 & 99.7 & 89.4 & 99.7 & 89.7 & 99.7 & 89.6  \\
     & & Ours     & 27.7 & 20.8 & 25.3 & 30.9 & 36.4 & 38.8 & 95.1 & 24.6 & 79.6 & 36.9  \\
     \cline{1-13}
     \multirow{8}{*}{DFDCP}
     & \multirow{2}{*}{SimSwap}
     & Random   & 99.4 & 86.6 & 97.4 & 84.9 & 98.3 & 85.9 & 97.7 & 86.1 & 98.1 & 86.8  \\
     & & Ours     & 80.5 & 12.2 & 56.4 & 16.3 & 88.5 & 35.0 & 89.2 & 25.1 & 84.4 & 32.2  \\
     \cline{2-13}
     & \multirow{2}{*}{InfoSwap}
     & Random   & 98.2 & 89.1 & 96.2 & 88.0 & 97.1 & 89.1 & 96.3 & 88.0 & 96.8 & 89.4  \\
     & & Ours     & 76.3 & 20.7 & 53.5 & 24.3 & 84.3 & 50.7 & 85.3 & 37.3 & 79.6 & 47.2  \\
     \cline{2-13}
     & \multirow{2}{*}{MobileFaceSwap}
     & Random   & 99.6 & 87.8 & 97.0 & 86.3 & 99.0 & 87.5 & 97.7 & 86.7 & 98.0 & 88.2  \\
     & & Ours     & 82.8 & 13.5 & 58.8 & 16.5 & 90.3 & 36.4 & 91.8 & 25.7 & 86.2 & 34.6  \\
     \cline{2-13}
     & \multirow{2}{*}{BlendSwap}
     & Random   & 99.4 & 86.1 & 97.2 & 84.8 & 98.4 & 85.7 & 97.3 & 85.6 & 97.8 & 85.8  \\
     & & Ours     & 84.8 & 14.3 & 59.5 & 18.2 & 91.8 & 37.8 & 93.9 & 26.9 & 87.4 & 34.5  \\
     \cline{1-13}
     \multirow{8}{*}{Avg.}
     & \multirow{2}{*}{SimSwap}
     & Random   & 99.7 & 88.7 & 99.6 & 87.9 & 99.4 & 88.4 & 99.2 & 88.3 & 99.3 & 88.5  \\
     & & Ours     & 63.1 & 13.8 & 41.9 & 22.2 & 72.5 & 31.0 & 89.7 & 22.8 & 78.1 & 27.0  \\
     \cline{2-13}
     & \multirow{2}{*}{InfoSwap}
     & Random   & 98.4 & 91.0 & 97.9 & 90.5 & 98.1 & 91.0 & 97.9 & 90.6 & 97.0 & 91.0  \\
     & & Ours     & 59.7 & 24.5 & 39.8 & 31.2 & 68.7 & 45.8 & 85.6 & 35.9 & 74.2 & 39.2  \\
     \cline{2-13}
     & \multirow{2}{*}{MobileFaceSwap}
     & Random   & 99.8 & 89.4 & 99.1 & 88.8 & 99.6 & 89.3 & 99.3 & 88.9 & 99.4 & 89.3  \\
     & & Ours     & 64.8 & 17.5 & 43.2 & 22.7 & 74.2 & 32.2 & 91.9 & 23.8 & 79.6 & 28.3  \\
     \cline{2-13}
     & \multirow{2}{*}{BlendSwap}
     & Random   & 99.6 & 87.9 & 99.1 & 87.3 & 99.3 & 87.8 & 99.1 & 87.5 & 99.2 & 87.6  \\
     & & Ours     & 66.6 & 16.5 & 43.9 & 23.6 & 76.0 & 33.7 & 94.3 & 25.2 & 80.6 & 29.1  \\
     \cline{1-13}
    \end{tabular}
\end{table*}

\begin{table}[!ht]
    \centering
    \caption{\small Obstructing inference: SSIM score ($\%$) of synthesized faces by obstructing the inference phase of DeepFake models using different face detectors on the Celeb-DF dataset. }
    \vspace{-0.3cm}
    \label{tab:ssim_inference_faceswap}
    \begin{tabular}{c|c|c|c|c|c|c}
    \hline
    DF model & Method & RF & YF & PB & S3FD & DSFD  \\
    \hline
    \multirow{2}{*}{Origin}
    & Random    & 96.5 & 96.4 & 97.1 & 96.1 & 97.1  \\
    & Ours      & 11.6 & 11.4 & 19.6 & 19.0 & 23.5  \\
    \hline
    \multirow{2}{*}{IAE}
    & Random    & 96.3 & 96.2 & 96.9 & 95.8 & 96.9  \\
    & Ours      & 11.8 & 11.3 & 19.3 & 18.7 & 23.3  \\
    \hline
    \multirow{2}{*}{LightWeight}
    & Random    & 96.4 & 96.2 & 97.0 & 95.9 & 97.0  \\
    & Ours      & 11.2 & 11.0 & 19.0 & 18.0 & 23.0  \\
    \hline
    \multirow{2}{*}{DFaker}
    & Random    & 96.1 & 95.9 & 96.7 & 95.7 & 96.5  \\
    & Ours      & 16.7 & 17.7 & 29.5 & 28.8 & 31.2  \\
    \hline
    \multirow{2}{*}{DFLH}
    & Random    & 96.0 & 95.9 & 96.5 & 95.5 & 96.4  \\
    & Ours      & 17.1 & 17.0 & 28.6 & 27.3 & 30.4  \\
    \hline
    \multirow{2}{*}{CDFv1}
    & Random    & 96.5 & 96.3 & 97.0 & 96.0 & 96.9  \\
    & Ours      & 17.9 & 17.7 & 30.5 & 28.7 & 31.7  \\
    \hline
    \multirow{2}{*}{CDFv2}
    & Random    & 96.9 & 96.8 & 97.3 & 96.7 & 97.3  \\
    & Ours      & 27.1 & 24.1 & 41.8 & 40.4 & 41.9  \\
    \hline
    \end{tabular}

\end{table}

\subsection{Results of Obstructing DeepFake Generation}

\subsubsection{Obstructing Inference}
We validate the effectiveness of our method by obstructing the inference process of all DeepFake models on FF++, Celeb-DF, DFDC, and DFDCP datasets with all face detectors. For each dataset, we randomly select three identities (id0, id1, id2) and make three pairs of identities. Then we contaminate one identity in each pair using our method and randomly select a detection box as the input face. We use an empty image (all 0 pixels) as the input face for the cases that contain no detection boxes.

\begin{figure}[!t]
\centering
\includegraphics[width=\linewidth]{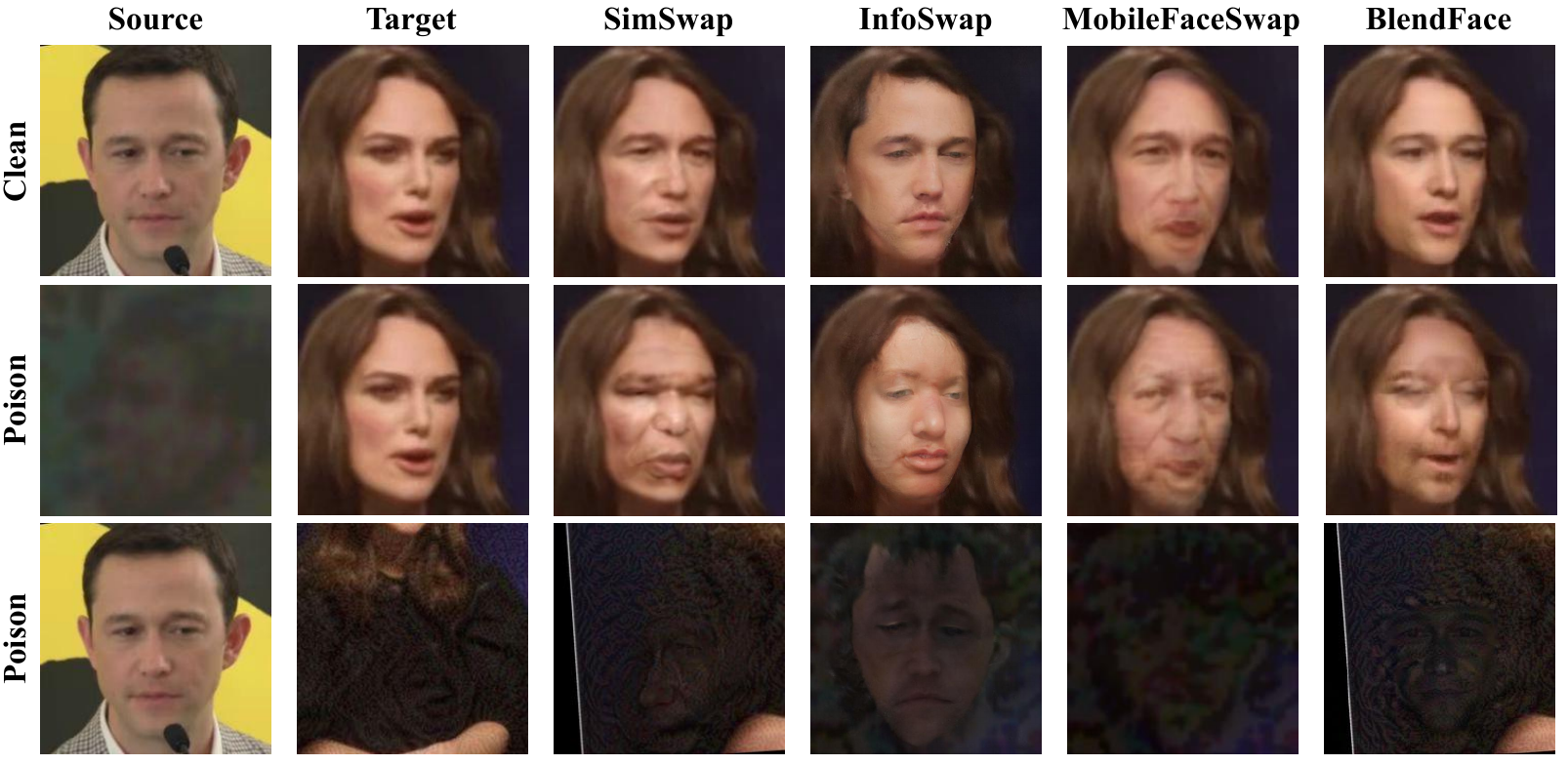}
\vspace{-0.7cm}
\caption{\small Visual examples of our method obstructing the inference phase of DeepFake models. The first row shows the effect of results using clean images. The second and third rows show the results corresponding to AO (source image is polluted while target image is original) and OA (source image is original while target image is polluted) scenarios. It can be seen that the polluted image can not extract the face correctly, thus notably degrading the visual quality of generated faces.}
\label{fig:vis_inference}
\end{figure}

For the DeepFake models SimSwap, InfoSwap, MobileFaceSwap, and BlendSwap, we use their officially released weights and generate faces with default settings. The performance of obstructing inference is shown in Table~\ref{tab:ssim_inference}. Since these models require both source and target identity as input, we design two scenarios to fully demonstrate the efficacy of our method. The first is ``AO'', where the source identity is polluted while the target identity is original. The second is ``OA'', which represents the opposite scenario. ``Random'' represents a baseline where random Gaussian noise is added instead of adversarial perturbations, with equivalent distortion strength. The synthesis quality is assessed using SSIM, where higher scores indicate better visual quality. The results show that random noise can hardly affect the synthesis quality, whereas our method achieves favorable performance under both scenarios, notably degrading the visual quality of DeepFake output. Fig.~\ref{fig:vis_inference} shows examples of our method obstructing these DeepFake models under AO and OA scenarios.

Note that in the inference phase, DeepFake models such as Origin, IAE, LightWeight, DFaker, DFLH, CDFv1, and CDFv2 require the same identity used in the training phase. Therefore, we infer three variants of each model using the corresponding images of all identity pairs on the Celeb-DF dataset. Table~\ref{tab:ssim_inference_faceswap} shows corresponding results, revealing that our method greatly reduces the visual quality of DeepFake faces.

\begin{table}[!t]
    \centering
    \caption{\small Obstructing training: SSIM score ($\%$) of synthesized faces using our method on the Celeb-DF dataset.}
    \vspace{-0.3cm}
    \label{tab:ssim}
    \begin{tabular}{c|c|c|c|c|c|c}
    \hline
    DF model & Method & RF & YF & PB & S3FD & DSFD  \\
    \hline
    \multirow{2}{*}{Origin}
    & Random    & 90.8 & 90.5 & 90.8 & 90.6 & 90.7  \\
    & Ours      & 62.9 & 70.4 & 75.7 & 76.7 & 83.9  \\
    \hline
    \multirow{2}{*}{IAE}
    & Random    & 92.2 & 92.1 & 92.1 & 92.1 & 92.3  \\
    & Ours      & 48.5 & 1.7 & 80.3 & 82.5 & 86.6  \\
    \hline
    \multirow{2}{*}{LightWeight}
    & Random    & 92.3 & 92.4 & 92.4 & 92.3 & 92.2  \\
    & Ours      & 71.4 & 77.0 & 80.9 & 81.9 & 86.9  \\
    \hline
    \multirow{2}{*}{DFaker}
    & Random    & 89.3 & 89.6 & 89.9 & 89.7 & 89.3  \\
    & Ours      & 41.9 & 2.2 & 70.6 & 72.5 & 56.7  \\
    \hline
    \multirow{2}{*}{DFLH}
    & Random    & 93.2 & 92.9 & 92.9 & 92.8 & 92.9  \\
    & Ours      & 49.4 & 2.4 & 80.4 & 81.9 & 87.1  \\
    \hline
    \multirow{2}{*}{CDFv1}
    & Random    & 88.7 & 88.9 & 88.7 & 88.7 & 88.9  \\
    & Ours      & 42.9 & 25.1 & 73.4 & 74.1 & 81.5  \\
    \hline
    \multirow{2}{*}{CDFv2}
    & Random    & 91.0 & 90.6 & 90.8 & 90.6 & 91.2  \\
    & Ours      & 48.9 & 3.2 & 79.7 & 80.5 & 85.0  \\
    \hline
    \end{tabular}
\end{table}

\subsubsection{Obstructing Training}
To validate the effectiveness of obstructing training, we apply our method to pollute the training faces of one identity while leaving the other one untouched. Following a similar setup as in the previous section, we select three identities (id0, id1, id2) and train DeepFake models using these identity pairs. Our method is validated using all face detectors. For each detector, we train $7 \times 3 \times 3 = 63$ models: seven DeepFake models trained individually three times, corresponding to three identity pairs, with each pair having three variants—clean training data, training data with random noise, and training data polluted by our method. Since our method is validated on five face detectors, a total of $\bm{63 \times 5 =315}$ DeepFake models are trained.

\begin{figure}[!t]
\centering
\includegraphics[width=1.0\linewidth]{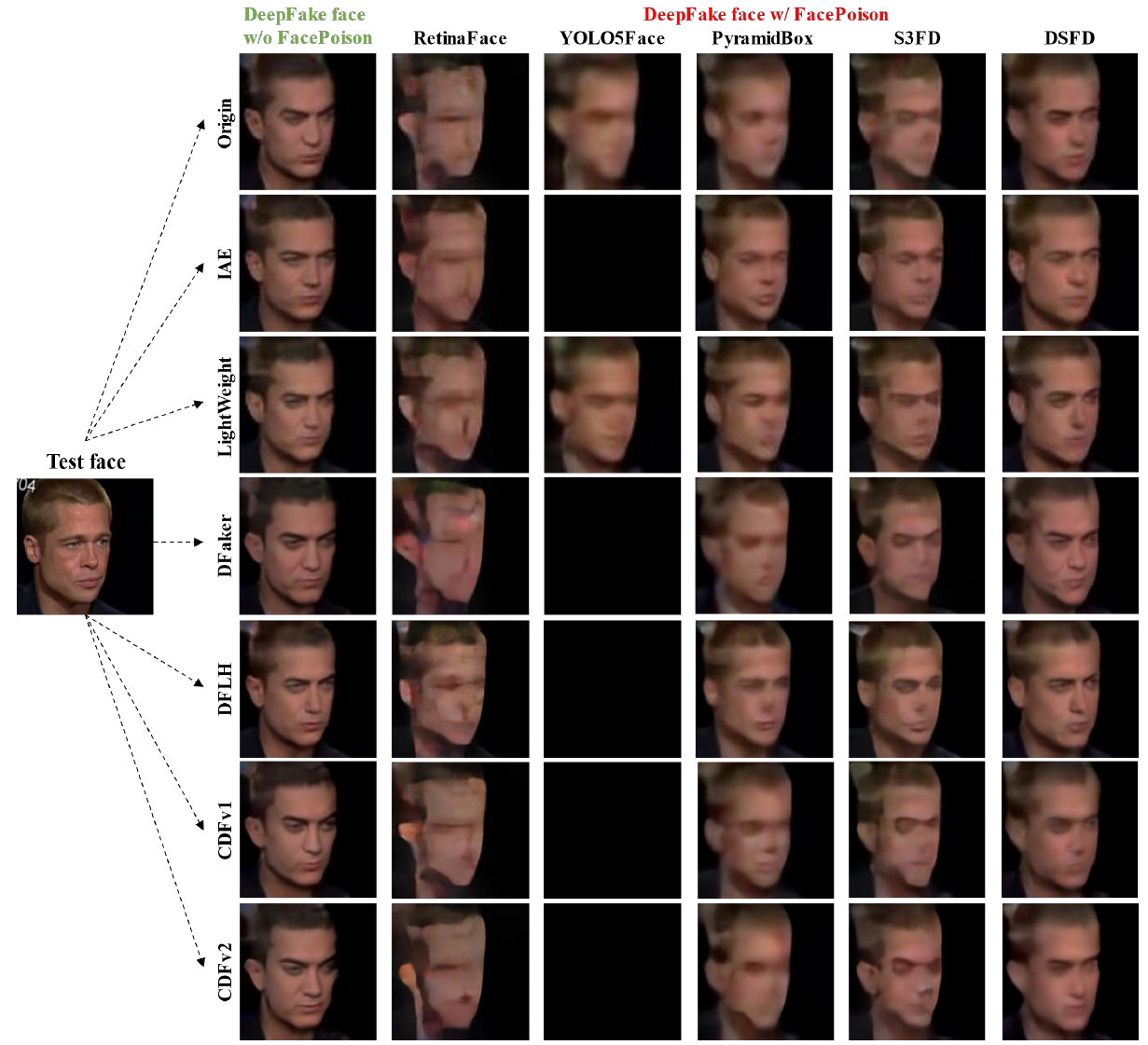}
\vspace{-0.7cm}
\caption{\small Visual examples of DeepFake faces. The first column shows DeepFake faces without our method. The other right columns show DeepFake faces with our method.} 
\label{fig:examples}
\end{figure}

After training, we feed correct face images into obstructed DeepFake models to see whether the synthesized faces are significantly disrupted. 
Table \ref{tab:ssim} shows the SSIM score of synthesized faces after using our method. We can observe that the random noise can hardly affect the synthesis quality, but our method can greatly reduce the SSIM score, demonstrating the efficacy of our method in obstructing the training of DeepFake models. Fig.~\ref{fig:examples} shows visual examples. It can be seen that our method can effectively degrade the visual quality of DeepFake faces. 

Remarkably, our method reduces the SSIM scores to single digits for models trained with the YOLO5Face detector. This is because our method fully disrupts this face detector, resulting in no detection boxes appearing. Consequently, no areas are extracted as training faces, leaving the corresponding regions as all-zero pixels. Therefore, given an input face, this DeepFake model can only generate images with all-zero pixels, as shown in Fig.~\ref{fig:examples}.

Training a single model is time-consuming, typically taking around, making it very difficult to validate all identities in Celeb-DF. To verify the generalizability of our method to different identity selections, we randomly select three additional sets of identities: (id2, id4, id10), (id12, id23, id5), and (id6, id9, id16). Using the PyramidBox face detector, we train the Origin DeepFake model with these identity sets. The results are shown in Table ~\ref{tab:ssim_more}, which reveals that show that identity selection does not affect the effectiveness of our method.

\begin{table}[t]
    \centering
    \caption{\small Obstructing training: SSIM score ($\%$) of synthesized faces using PyramidBox face detector on other identities.}
    \vspace{-0.3cm}
    \label{tab:ssim_more}
    \begin{tabular}{c|ccc}
    \hline
    Origin, PB & (id2, id4, id10) & (id12, id23, id5) & (id6, id9, id16) \\
    \hline 
    Random   & 92.4 & 90.1 & 90.8 \\ 
    Ours     & 74.8 & 73.6 & 71.8 \\
    \hline
    \end{tabular}
\end{table}

Moreover, we perform our method on FF++ datasets for a comprehensive evaluation. We randomly select three identities, extract faces using the PyramidBox detector, and train all DeepFake models following the same protocol. The performance is shown in Table~\ref{tab:ssim_ff}, demonstrating the effectiveness of our method in disrupting the generative quality. 

\begin{table}[t]
    \centering
    \small
    \caption{\small Obstructing training: SSIM score ($\%$) of synthesized faces using our method on FF++ dataset.}
    \vspace{-0.3cm}
    \label{tab:ssim_ff}
    \begin{tabular}{c|cc}
    \hline
    DF model & Random & Ours  \\
    \hline 
    Origin & 81.7 & 32.7 \\
    \hline
    IAE & 79.5 & 49.1 \\
    \hline
    LightWeight & 82.8 & 36.7 \\
    \hline
    DFaker & 84.3 & 38.8 \\
    \hline
    DFLH & 84.6 & 50.8 \\
    \hline
    CDFv1 & 79.2 & 42.6 \\
    \hline
    CDFv2 & 85.1 & 58.8 \\
    \hline
    \end{tabular}
\end{table}

\subsubsection{Effect of Poison Ratio}
\begin{figure}[!t]
    \centering
    \hspace{-1cm}\includegraphics[width=0.6\linewidth]{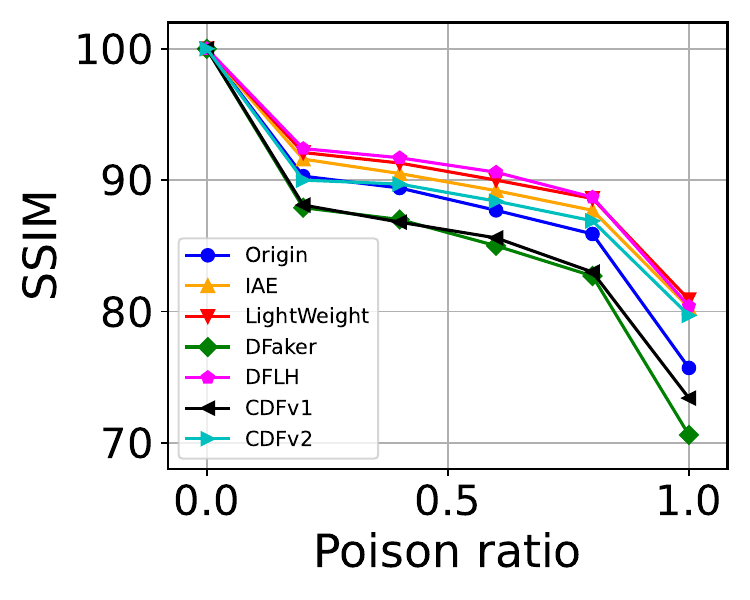}
    \vspace{-0.3cm}
    \caption{\small Effect of different poison ratios used for polluting training faces. $0$ denotes all training samples are clean and $1$ denotes our method is applied to all training faces.}
    \label{fig:ratio}
\end{figure}

This part investigates the effect of the poison ratio of training faces, \ie, the relationship between the synthesis quality and the ratio of incorrect faces in training. Specifically, we change the poison ratio in a range of $[0, 1]$, where $0$ denotes our method is not applied and $1$ denotes our method is applied to all training faces. Fig. \ref{fig:ratio} illustrates the curve of the SSIM score with different poison ratios. It can be seen that the training faces do not need to be fully disrupted by our method, in order to reduce the synthesis quality, \eg, the SSIM score is still around $90\%$ at poison ratio $0.2$.

\subsubsection{Adapted Comparisons}
For a comprehensive evaluation, we incorporate two representative works, CMUA~\cite{huang2022cmua} and DF-RAP~\cite{qu2024df}, as baselines and assess their performance of obstructing inference on the Celeb-DF dataset. Note that both methods are designed to directly disrupt DeepFake output rather than interfere with the face detection process, thus not fully aligning with our setting. For a fair comparison, we follow their default setting and adapt them to the extracted face images. For CMUA, we use the universal perturbation provided in their official repository. For DF-RAP, we utilize the pretrained generator they released. As shown in Table~\ref{tab:ssim_inference_baseline}, CMUA and DF-RAP exhibit unsatisfactory effectiveness in disrupting the synthesized results when applied to DeepFake models outside of those used in their original studies. This indicates their limited generalizability, in contrast to the broader applicability demonstrated by our defense strategy. 

\begin{table*}[!ht]
    \centering
    \caption{\small Adapted comparisons with others. }
    \vspace{-0.3cm}
    \label{tab:ssim_inference_baseline}
    \begin{tabular}{c|c|cc|cc|cc|cc|cc}
    \hline
    \multirow{2}{*}{DF model}
    & \multirow{2}{*}{Method}
    & \multicolumn{2}{c|}{RetinaFace}
    & \multicolumn{2}{c|}{YOLO5Face}
    & \multicolumn{2}{c|}{PyramidBox}
    & \multicolumn{2}{c|}{S3FD}
    & \multicolumn{2}{c}{DSFD} \\
    \cline{3-12}
    & & AO & OA & AO & OA & AO & OA & AO & OA & AO & OA \\
    \hline
     \multirow{3}{*}{SimSwap}
     & CMUA   & 99.6 & 87.9 & 99.5 & 87.8 & 99.5 & 87.8 & 99.4 & 87.8 & 99.5 & 87.9  \\
     & DF-RAP   & 99.7 & 91.0 & 99.6 & 91.0 & 99.6 & 91.0 & 99.6 & 91.0 & 99.7 & 91.0  \\
     & Ours     & 74.0 & 14.1 & 47.5 & 18.7 & 82.6 & 28.3 & 88.9 & 27.6 & 66.8 & 33.4  \\
     \hline
     \multirow{3}{*}{InfoSwap}
     & CMUA   & 98.6 & 94.6 & 99.5 & 94.6 & 98.4 & 94.5 & 98.4 & 94.5 & 98.5 & 94.6  \\
     & DF-RAP   & 98.6 & 97.0 & 99.7 & 97.0 & 98.5 & 96.9 & 98.5 & 96.9 & 98.6 & 97.0  \\
     & Ours     & 70.3 & 28.6 & 44.9 & 28.5 & 77.9 & 43.6 & 84.1 & 43.6 & 63.2 & 48.5  \\
     \hline
     \multirow{3}{*}{MobileFaceSwap}
     & CMUA   & 99.8 & 88.6 & 98.5 & 88.6 & 99.7 & 88.6 & 99.7 & 88.5 & 99.8 & 88.6  \\
     & DF-RAP   & 99.9 & 92.5 & 98.6 & 92.4 & 99.8 & 92.4 & 99.9 & 92.5 & 99.9 & 92.5  \\
     & Ours     & 76.4 & 15.3 & 49.2 & 19.2 & 85.2 & 29.5 & 91.7 & 29.0 & 68.8 & 35.0  \\
     \hline
     \multirow{3}{*}{BlendSwap}
     & CMUA   & 99.5 & 76.4 & 99.8 & 76.2 & 99.5 & 76.2 & 99.4 & 76.1 & 99.5 & 76.2  \\
     & DF-RAP   & 99.6 & 75.7 & 99.9 & 75.7 & 99.6 & 75.7 & 99.6 & 75.7 & 99.6 & 75.7  \\
     & Ours     & 78.7 & 18.4 & 50.2 & 20.1 & 87.1 & 31.0 & 93.8 & 30.3 & 70.0 & 36.1  \\

     \hline
    \end{tabular}
\end{table*}


\subsection{Discussion}
\label{sec:discussion}



\subsubsection{F1-score VS Average Precision (AP)}
\begin{figure}[!ht]
\centering
\includegraphics[width=\linewidth]{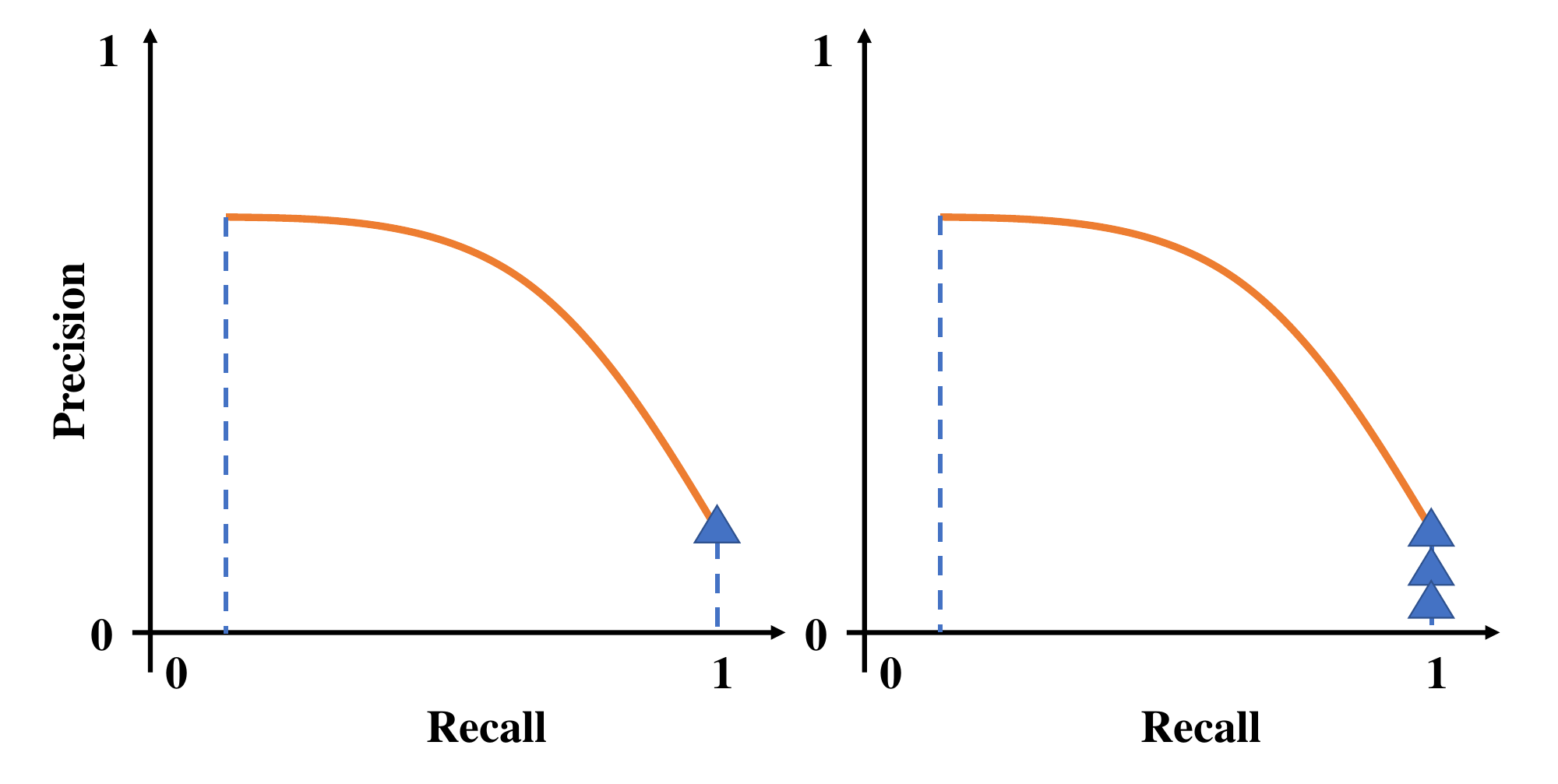}
\vspace{-0.7cm}
\caption{\small Comparison of PR curves in different detection scenarios. The left example has a moderate number of candidate proposals (blue triangles), while the right has significantly more. Despite both detecting all ground-truth faces (recall = 1), the excessive redundant proposals in the right example reduce precision. However, the area under the curve (AP) remains unchanged. This illustrates that AP is not sensitive to redundancy. In contrast, our proposed use of F1-score penalizes such redundant detections, providing a more accurate assessment for defense purposes.}
\label{fig:pr}
\end{figure}

In the context of face detection, candidate proposals refer to all possible face bounding boxes predicted by the detector, each associated with a different confidence score. These proposals are essential for computing the Average Precision (AP) score, a widely used metric in the evaluation of face detection. 
To calculate the AP, all candidate proposals are first sorted by their confidence scores. Then, precision and recall are computed at various thresholds, where precision indicates the proportion of predicted boxes that are accurate, and recall reflects the proportion of ground-truth faces that are correctly detected. The AP is obtained by plotting the precision-recall curve and calculating the area under this curve. 
However, this metric is not sensitive to redundant face candidate proposals, see Fig.~\ref{fig:pr}. In face detection, this insensitivity is acceptable: high Average Precision (AP) can still be achieved as long as all faces are correctly detected, even if many redundant proposals exist. But it becomes problematic in our task, where redundant face proposals can significantly contaminate the training face set. Therefore, we use the F1-score to directly measure precision and recall, providing a more appropriate and reliable evaluation metric for our specific objective.

\subsubsection{Limitations} Our method is specifically dedicated to defending face-swap based DeepFake, which relies on extracting the face region as a critical initial step. Therefore, it is not effective for other DeepFake techniques mentioned in related works, such as face editing and entire-face generation, as face detection is not the prerequisite step in their inference or training process. As a potential direction for future work, it would be valuable to explore hybrid defense strategies that not only disrupt the face detection pipeline but also interfere with latent facial attribute representations within generative models, thereby improving the generalizability of proactive defense mechanisms.

Another limitation is that our method focuses on the modern DNN-based face detectors. Traditional face detection tools, such as those based on non-DNN machine learning models (\eg, SVMs in Dlib \cite{dlib09}), are difficult to be affected by our method. These non-DNN models typically have far fewer parameters than DNN models, making it more challenging to create effective adversarial perturbations. In future work, we intend to explore more advanced strategies beyond adversarial perturbation, such as manipulating high-level semantic presentations of facial appearance. This may help bypass architectural constraints and extend defense capabilities across diverse detector types. While our current method targets DNN-based face detectors, it serves as a proof of concept that proactive DeepFake defense via disrupting the face detection pipeline is both feasible and impactful, offering fresh insight for the community of DeepFake defense.

Furthermore, our method may be less effective against adaptive countermeasures employed by forgery makers, as discussed in Sec.~\ref{sec:robustness}. It is thus our continuing effort to improve the robustness of the adversarial effectiveness.

\section{Conclusion}
DeepFake is becoming a problem that is encroaching on our trust in online media. As DeepFake requires automatic face detection as an indispensable pre-processing step in preparing training data, an effective protection scheme can be obtained by disrupting the face detection methods. In this work, we develop a proactive protection method (FacePoison) to deter bulk reuse of automatically detected faces for the production of DeepFake faces. Our method exploits the sensitivity of DNN-based face detectors and uses adversarial perturbation to contaminate the face sets. This is achieved by a dedicated adaptation of mainstream adversarial attacks to disrupt face detectors.
Moreover, we introduce VideoFacePoison, an extended strategy to propagate the FacePoison across frames. In contrast to applying FacePoison to each frame individually, VideoFacePoison can only calculate adversarial perturbations on a certain set of frames and estimate the perturbations for others, while retaining the effectiveness of disrupting face detection. The experiments are conducted on eleven different DeepFake models with five different face detectors and empirically show the effectiveness of our method in obstructing DeepFakes in both inference and training phases.

\textit{On one hand}, we expect this technology can offer fresh insights for the proactive DeepFake defense research and inspire more effective solutions to break the pipeline of the DeepFake generation, instead of only focusing on disrupting the DeepFake models themselves. \textit{On the other hand}, the current forensics strategies are impossibly perfect for arbitrary scenarios, and these strategies can be hacked with a high probability once they are released. \textit{Nevertheless, they still play a significant role. Even when hacked, these strategies raise the bar for creating forgeries, thereby slowing down the pace of their creation and dissemination.}

\smallskip
\noindent{\bf Acknowledgement.} This work is supported in part by the National Natural Science Foundation of China (No.62402464), Shandong Natural Science Foundation (No.ZR2024QF035), China Postdoctoral Science Foundation (No.2021TQ0314, No.2021M703036), and Sanya Science and Technology Special Fund (No.2022KJCX92). Jiaran Zhou is supported by the National Natural Science Foundation of China (No.62102380) and Shandong Natural Science Foundation (No.ZR2021QF095, No.ZR2024MF083). Baoyuan Wu is supported by Guangdong Basic and Applied Basic Research Foundation (No. 2024B1515020095), National Natural Science Foundation of China (No. 62076213), Shenzhen Science and Technology Program under grants (No. RCYX20210609103057050).

\bibliographystyle{IEEEtran}
\bibliography{ref}

\end{document}